\documentclass[preprint,10pt]{elsarticle}

\usepackage[letterpaper,margin=1in]{geometry}

\usepackage{amssymb}
\usepackage{amsmath}
\usepackage{algorithmic}
\usepackage{algorithm}
\usepackage{subcaption} 
\usepackage{changepage}
\usepackage{float}
\usepackage{bm}
\usepackage{booktabs}
\usepackage{multirow}
\usepackage{tabularx}
\usepackage{hyperref}
\usepackage{xcolor}
\usepackage{makecell}
\definecolor{darkgray}{gray}{0.4}

\begin{document}

\begin{frontmatter}

\title{FedFFT: Taming Client Drift in Federated SAM via Spectral Perturbation Filtering}





\author[addr1]{Liyang Yuan\fnref{grant}}
\ead{yuanly24@mails.jlu.edu.cn}

\author[addr2]{Yibo Yang}
\ead{yibo.yang93@gmail.com}

\author[addr1]{Dandan Guo\corref{cor1}\fnref{grant}}
\ead{guodandan@jlu.edu.cn}

\cortext[cor1]{Corresponding author}

\fntext[grant]{This work of Liyang Yuan and Dandan Guo was supported by the National Natural Science Foundation of China (NSFC) under Grant No. 62306125.}

\affiliation[addr1]{organization={School of Artificial Intelligence, Jilin University},
            city={Changchun},
            country={China}}

\affiliation[addr2]{organization={King Abdullah University of Science and Technology},
            country={Saudi Arabia}}

\begin{abstract}
Federated Learning (FL) enables decentralized training without data sharing, but suffers from statistical heterogeneity across clients, leading to client drift, poor generalization, and sharp minima compared to centralized training. Sharpness-Aware Minimization (SAM) has emerged as a promising approach to improve generalization, yet its application in federated learning still suffers from divergence problems, since perturbations are computed locally and reflect client-specific loss geometries. To better understand this issue, {we provide experimental evidence from a new perspective, the frequency domain, for SAM perturbations in federated settings, revealing that inter-client perturbation inconsistencies are predominantly concentrated in the low-frequency spectrum.}
Motivated by this insight, we propose \textbf{Fed}erated learning with \textbf{F}requency-domain \textbf{F}iltering of SAM per\textbf{t}urbations (\textbf{FedFFT}). It is a lightweight and plug-and-play method that filters out low-frequency components of SAM perturbations without requiring additional communication, thereby suppressing inconsistent components in client updates while preserving consistent learning signals. Extensive experiments across multiple benchmarks and diverse backbones demonstrate that FedFFT consistently outperforms SAM-based FL methods, particularly under severe non-IID distributions. These results highlight the effectiveness, scalability, and general applicability of our frequency-domain perspective for sharpness-aware federated optimization.
\end{abstract}




\end{frontmatter}
\setcounter{section}{0}
\section{Introduction}
\label{sec:intro}

{Federated Learning (FL) \citep{fedavg} is a distributed learning paradigm where multiple clients collaboratively train a global model under the coordination of a central server, while keeping their raw data local to preserve privacy. In each communication round, clients perform local training and only transmit model parameters or updates, which are aggregated to update the global model. This framework has been widely applied in privacy-sensitive domains such as healthcare, finance, mobile applications, and autonomous systems \citep{fl_apply1,fl_apply2,fl_apply3,pr-survey}. However, the practical effectiveness of FL is severely hampered by the statistical heterogeneity inherent in real-world data, where the local data distributions across clients are typically non-independent and identically distributed (Non-IID). This causes the optimization objectives of individual clients to become misaligned with one another and with the global goal, resulting in local updates that pull the shared model in conflicting directions. This phenomenon are known as client drift \citep{scaffold,fl_challenge1,fl_challenge2,fl_challenge3, fednn}, which not only slows convergence but can also limits the model's ability to generalize to the overall underlying distribution. }

{To address the generalization challenges posed by client drift, a promising research direction has shifted from traditional client-side regularization \citep{feddyn,scaffold,fedprox,moon,fedcm} or aggregation methods \citep{feddisco,fedlaw,fedawa} to exploring the geometry of the loss landscape. These methods build upon the insight that convergence to sharp minima correlates with poor generalization \citep{fl_challenge2}, where flatter minima often yield better performance. Sharpness-Aware Minimization (SAM) \citep{sam} is a representative technique designed for this purpose, seeking flatter regions by optimizing a perturbed loss. Building on this, FedSAM \citep{qusam,caldarola2022improving} pioneered the application of SAM to local training in Federated Learning. While FedSAM \citep{fedsmoo, fedlesam} have shown strong performance across different settings, they primarily focus on local flatness, implicitly assuming that minimizing sharpness locally leads to a globally flat minimum. In practice, however, under substantial data heterogeneity, the local and global loss landscapes may diverge considerably, and improvements in local flatness do not necessarily guarantee global flatness. Subsequent works \citep{fedsmoo,fedlesam,fedgloss,fedgamma} have attempted to bridge this local-global gap through various strategies, such as enhancing client-side updates \citep{fedsmoo} or shifting the sharpness optimization to the server \citep{fedgloss}, sometimes employing complex frameworks like the Alternating Direction Method of Multipliers (ADMM) \citep{admm}. Despite these advances, existing solutions face a difficult trade-off: purely client-side methods struggle with the deviation of local sharpness estimates, while server-centric approaches often come at the cost of significant communication or computational overhead. Crucially, none of these approaches explicitly investigate the intrinsic nature of the perturbations themselves, leaving open the question of whether they can be refined to better align clients with the global objective. 
}

{In this work, we address this question by introducing a novel frequency-domain perspective. 
{To our knowledge, we are the first to conduct a systematic experimental study on the spectral properties of client-side SAM perturbations in FL. } Our key finding is that the inter-client disagreements are not random noise; they are predominantly concentrated in the low-frequency spectrum, which we hypothesize is strongly tied to client-specific data biases. Motivated by this insight, we propose Federated learning with Frequency-domain Filtering of SAM perturbations (FedFFT), a lightweight and plug-and-play method. FedFFT applies a high-pass filter to the locally computed perturbations, systematically removing the discordant low-frequency components while preserving  consistent learning signals in the higher frequencies. This alignment in the spectral domain is achieved without requiring any additional communication overhead. Notably, FedFFT can be seamlessly integrated as a plug-and-play module with  federated learning framework that utilizes client-side SAM optimizers, further broadening its applicability. Our contributions are summarized as below:} \textbf{(1) Frequency-domain analysis of perturbations.} We provide the first study of SAM perturbations in FL across clients based on spectral decomposition, and reveal that heterogeneity is primarily concentrated in the low-frequency bands. \textbf{(2) Algorithm design.} Based on this insight, we introduce {FedFFT}, a simple yet effective approach that filters out low-frequency perturbation components to suppress inconsistent client updates while retaining retaining consistent learning signals. \textbf{(3) Extensive empirical validation.} We conduct comprehensive experiments across multiple benchmarks and backbones, under varying degrees of data heterogeneity. The results show that FedFFT outperforms related baselines, particularly in highly non-IID settings, demonstrating both effectiveness and scalability.

\section{Related work}
\label{sec:related}

In this section, we provide an overview of the research most relevant to our work. 
Specifically, we first review the Sharpness-Aware Minimization (SAM) methodology and its various extensions in deep learning. 
Next, we summarize key developments in Federated Learning (FL), highlighting the challenges arising from statistical heterogeneity and client drift. 
Finally, we discuss recent attempts to integrate SAM into FL, emphasizing how existing methods address generalization and convergence issues in decentralized settings.

\subsection{Sharpness-Aware Minimization (SAM)}

The connection between generalization and flat minima was first recognized in early studies \cite{flat_minima}, and later work confirmed that smoother loss landscapes generally lead to better generalization \cite{keskar2017large,neyshabur}. 
Building on this insight, Sharpness-Aware Minimization (SAM) was introduced as a PAC-Bayesian inspired method that explicitly minimizes loss sharpness and achieves strong generalization across image classification benchmarks \citep{sam}. 
Since then, numerous extensions have been developed. A scale-invariant version improves training stability \cite{kwon2021asam}, while another reformulates sharpness from both theoretical and intuitive perspectives \cite{zhuang2022}. 
Further studies focus on perturbation strategies, including adaptive or random amplitudes \cite{liu2022random,ahn2024}, dynamic adjustment through DSAM \cite{chen2024dsam}, and variance reduction across domains with DISAM \cite{2024disam}. 

\subsection{Federated Learning} 

Federated Learning (FL), introduced with FedAvg \cite{fedavg}, enables collaborative model training without raw data sharing. While preserving privacy, this decentralized design exacerbates the \emph{client-drift problem}—the divergence between local and global updates—mainly due to non-IID data and multi-step local errors \cite{feddyn,fl_challenge1,fl_challenge2}. Limited client participation further aggravates drift and degrades performance. To mitigate client drift, existing methods can be broadly grouped into two categories: (i) \textit{local objective regularization}, which modifies local training to align client updates with the global objective, such as SCAFFOLD \cite{scaffold} , FedProx \cite{fedprox} and FedDyn \cite{feddyn}; and (ii) \textit{modified aggregation strategies}, which design more robust global update rules beyond simple averaging, such as FedAWA \cite{fedawa}, FedLAW \cite{fedlaw}, and FedDisco \cite{feddisco}. While these approaches improve optimization stability, they are primarily rooted in empirical risk minimization and often overlook the relationship between the global loss landscape and generalization ability. This motivates a new  research line that leverages SAM in FL.  

\subsection{SAM in Federated Learning}

FedSAM \cite{qusam,caldarola2022improving} first brought SAM into FL by applying local perturbations to improve generalization.  Subsequent variants extended this idea.  For example, FedSpeed \cite{fedspeed} used an Alternating Direction Method of Multipliers (ADMM) framework to enhance communication efficiency. PLGU \cite{plgu} and FedSOL \cite{fedsol}, explored layer-wise perturbation and proximal-based corrections. FedGAMMA \cite{fedgamma} introduced variance-reduction techniques to align client updates from a global perspective.  FedSMOO \cite{fedsmoo} reduced inconsistency by correcting both updates and perturbations through ADMM, while FedLESAM \cite{fedlesam} introduced global perturbations to better guide local training. FedFSA \cite{fedfsa} focused on parameter sensitivity, applying stronger perturbations only to the most sensitive layers to balance convergence and generalization. FedGloSS \cite{fedgloss} shifted attention from local sharpness to global flatness by applying SAM on the server, highlighting the importance of global geometry in federated optimization. Unlike these approaches, our method is motivated by a novel spectral analysis, revealing inter-client disagreement in low-frequency components, which we filter to produce consistent perturbations and flatter minima. 

\subsection{Frequency Analysis in Deep Learning}
Frequency-based perspectives have been increasingly adopted to understand deep neural networks. Early studies revealed the spectral bias of neural networks, showing a tendency to learn low-frequency components first \cite{rahaman2018spectral, xu2019frequency}. Subsequent work leveraged frequency analysis to interpret network dynamics, such as characterizing generalization through Fourier-domain behaviors \cite{rahaman2018spectral}. Other methods focused on robustness, where perturbation analysis in the frequency domain helped explain vulnerability to high-frequency noise \cite{yin2019fourier,wang2019high}. Otherwise, the feasibility of treating parameter updates as signals is demonstrated by the work \cite{wang2018super}, which applies FFT to gradients for compression. This shows that optimization signals possess meaningful spectral structure.

Unlike these approaches, which analyze networks or inputs, our method conducts spectral analysis directly on SAM’s perturbation vector. We reveal that the perturbation contains structured low-frequency disagreement reflecting optimizer misalignment. By filtering these components, we produce more consistent perturbations that guide the optimization trajectory toward flatter minima.

\section{Background}
\label{sec:back}

\subsection{Sharpness-aware Minimization}
To improve model generalization and robustness, modern optimization methods have shifted focus from merely finding solutions with low training loss to finding solutions that reside in flat minima of the loss landscape. SAM~\citep{sam} is a leading technique for this purpose. It jointly minimizes the loss value and the sharpness by solving the following minimax objective:
\begin{equation}\label{eq:sam_objective}
    \min_{w} \max_{\|\delta\| \le \rho} \mathcal{L}(w + \delta),
\end{equation}
where $\mathcal{L}(\cdot)$ is the empirical loss on the training data, and $\rho$ is the neighborhood size. In practice, the inner maximization is approximated with a single step of gradient ascent. The full optimization process for parameter  $w$ involves two steps: (1) Compute the perturbation that approximately maximizes the loss:
    $ \delta^*(w) = \rho \frac{\nabla \mathcal{L}(w)}{\|\nabla \mathcal{L}(w)\|_2} $; (2) Update the model parameters using the gradient at the perturbed point:
    $ w \leftarrow w - \eta \nabla \mathcal{L}(w + \delta^*(w)) $. This procedure encourages the optimizer to converge to flat minima, which are empirically linked to better generalization performance.

\subsection{Federated Learning}
FL is a distributed learning paradigm that enables training a global model on data from $K$ clients, coordinated by a central server, without centralizing the private client datasets $\mathcal{D}_k$. The core objective in FL is to minimize the global empirical risk $F(w)$, defined as the weighted average of the local empirical losses $f_k(w)$:
\begin{equation}\label{eq:fl_objective}
    \min_{w } F(w) :=\frac{1}{K} \sum_{k=1}^{K}  f_k(w), \quad \text{where} \quad f_k(w) = \frac{1}{|\mathcal{D}_k|} \sum_{(x,y) \in \mathcal{D}_k} \mathcal{L}_k(w; x, y).
\end{equation}
The widely-adopted FedAvg algorithm~\citep{fedavg} solves this objective via iterative communication rounds. In each round, the server \textbf{(1) Broadcasts} the global model {$w^t$} to clients. Clients then perform \textbf{(2) Local Updates} on their data to produce $w_k^{t+1}$. These are \textbf{(3) Uploaded} to the server for \textbf {(4) Aggregation} 
into the new global model   {$w^{t+1} = \sum_k p_k w_k^{t+1}$.}

A key challenge in FL is {data heterogeneity}, where client data distributions are Non-Independent and Identically Distributed (Non-IID). This causes the local objectives $f_k(w)$ to be inconsistent with one another, leading to misaligned loss landscapes and the ``client drift'' phenomenon, which poses a significant challenge to training a robust global model.

\subsection{SAM in Federated Learning}
Given the challenge of training on misaligned local landscapes, a natural strategy is to seek solutions in flat minima. This motivates applying SAM to FL, known as FedSAM~\citep{qusam}, which incorporates SAM into each client's local training. The local and global objectives are:
\begin{equation}\label{eq:fedsam_objectives}
   \min_{w } F^{\text{SAM}}(w) := \frac{1}{K} \sum_{k=1}^{K}  f_k^{\text{SAM}}(w), \quad \text{where} \quad f_k^{\text{SAM}}(w) = \max_{\|\delta_k\| \le \rho} f_k(w + \delta_k).
\end{equation}
Compared with FedAvg, FedSAM differs in that it optimizes the sharpness-aware local objectives rather than the original local losses. While FedSAM encourages convergence to locally flat regions, it does not guarantee a flat global landscape. Under data heterogeneity, the locally computed perturbation vectors $\delta_k$, which capture the directions of sharpness, can themselves diverge. This raises a critical question that existing works have not fully explored: {what is the underlying structure of these inter-client perturbation disagreements?} Understanding this is key to mitigating their negative impact, which directly motivates our work.
\section{Methods}
\label{sec:methods}

\vspace{1em} 
\noindent\begin{minipage}{\textwidth}
    \centering
    \begin{minipage}{0.31\textwidth}
        \centering
        \includegraphics[width=\linewidth]{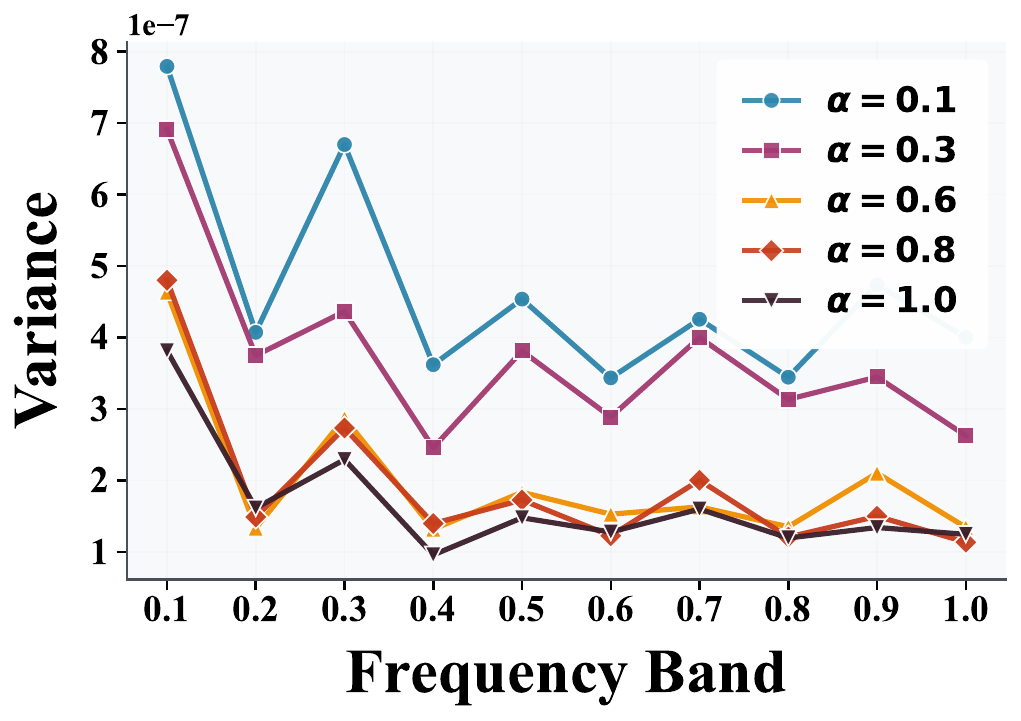}
        \centerline{\small (a) CIFAR-10}
    \end{minipage}\hfill
    \begin{minipage}{0.31\textwidth}
        \centering
        \includegraphics[width=\linewidth]{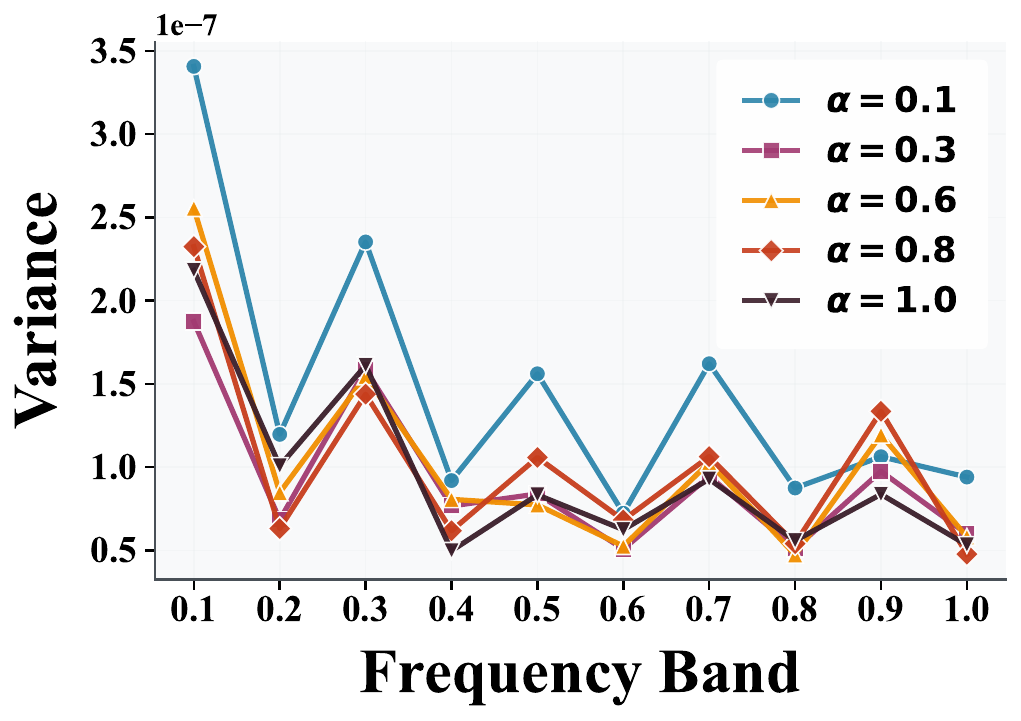}
        \centerline{\small (b) CIFAR-100}
    \end{minipage}\hfill
    \begin{minipage}{0.31\textwidth}
        \centering
        \includegraphics[width=\linewidth]{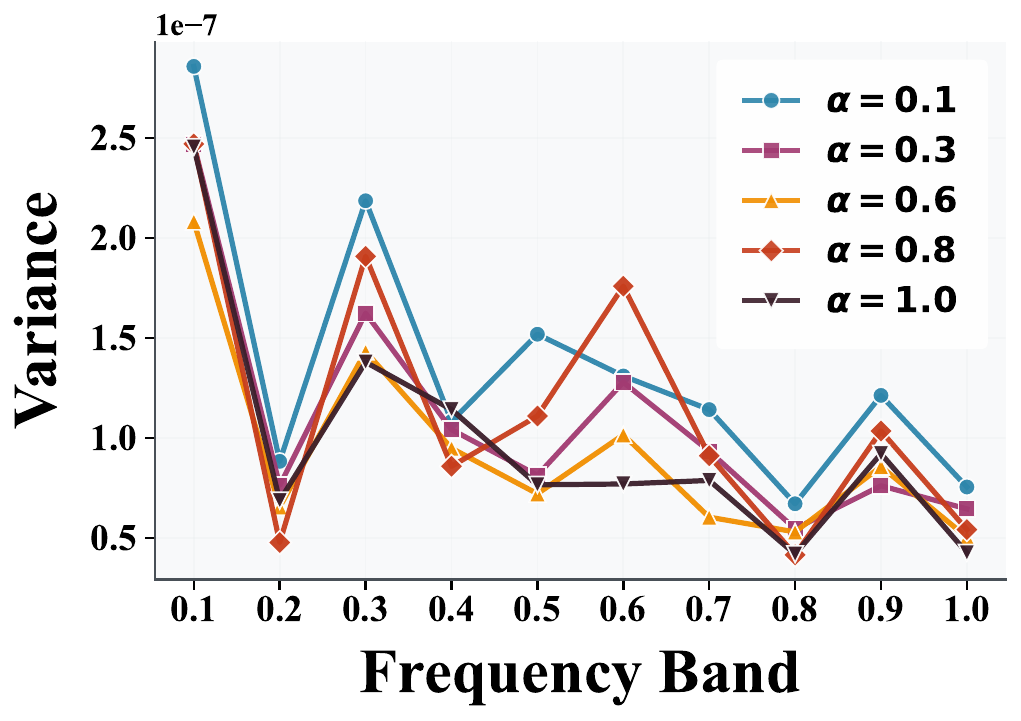}
        \centerline{\small (c) TinyImageNet}
    \end{minipage}
    
    \captionof{figure}{\small{(a–c) Variance of SAM perturbations across clients under different levels of data heterogeneity, controlled by the Dirichlet concentration parameter $\alpha$. Across all datasets and settings, inter-client variance is consistently concentrated in the low-frequency components, indicating that client disagreement is primarily a low-frequency phenomenon. }}
    \label{fig:frequency_analysis}
\end{minipage}
\vspace{1em} 

In this section, we introduce our proposed method, Federated learning with Frequency domain Filtering of SAM perturbations (FedFFT). We begin by elaborating on the motivation stemming from our observations of client perturbations in federated environments. Subsequently, we provide a detailed description of the FedFFT algorithm.

\subsection{ Motivation: Analyzing Client Perturbations in the Frequency Domain}

In FL, non-IID data distributions among clients are the primary cause of the client drift phenomenon. While Sharpness-Aware Minimization (SAM) is a powerful technique for improving generalization, its application in FL may amplify this issue. Since SAM perturbations are computed locally, they intrinsically reflect the geometry of client-specific loss landscapes, causing the perturbation vectors themselves to diverge. However, the underlying structure of these perturbation divergences remains poorly understood. To bridge this gap, we introduce a novel diagnostic approach. 

 To the best of our knowledge, this work is the first to employ frequency-domain analysis to systematically investigate the nature of SAM perturbation disagreements in a federated context. We treat each client's perturbation vector as a signal and use the Real-valued Fast Fourier Transform (RFFT) to observe its characteristics. Our analysis is conducted on a per-layer basis to respect the model's architectural integrity, with the full implementation details provided in below Section \ref{ref:method}. For each layer, we compute the variance across clients within different frequency bands. As illustrated in Figure \ref{fig:frequency_analysis}, which averages these variances across all layers, we discover a clear pattern: 
 
 \emph{The primary disagreements are concentrated in the low-frequency components, while the disagreements in high-frequency components are much smaller. This finding suggests that the low-frequency parts are highly correlated with client-specific biases, whereas the high-frequency parts may represent more consistent features of the shared learning task.}

Based on this key insight, we formulate our central hypothesis: by filtering out the discordant low-frequency components while preserving the consistent high-frequency information, we can mitigate client drift and improve the global model's performance.

\subsection{FedFFT: The Proposed Method}

Considering that inter-client disagreements are predominantly a low-frequency phenomenon, we propose \textbf{Fed}erated learning with \textbf{F}requency-domain \textbf{F}iltering of SAM per\textbf{t}urbations (\textbf{FedFFT}). Our method refines the client-side SAM update by integrating a frequency filtering module. This module is designed as a lightweight, plug-and-play replacement for the standard perturbation calculation within any federated learning framework that employs SAM-based optimizers on client devices. 

The central motivation for FedFFT stems from a key observation: the SAM perturbation vector itself behaves as an optimization signal with meaningful spectral structure. Earlier analyses focus primarily on the frequency properties of model parameters or input data, but recent systems work has demonstrated that optimization signals—such as gradients—also admit interpretable and structured frequency characteristics. Notably, FFT-based Gradient Sparsification \cite{wang2018super} applies the Fast Fourier Transform  to gradients and shows that a significant portion of their energy is concentrated in specific frequency bands, enabling principled compression. This establishes an important precedent: optimization vectors are not noise-like; they possess stable, compressible, and manipulable spectral patterns.

FedFFT builds upon this foundation but innovates in two crucial ways:
(1) We apply spectral analysis to SAM perturbations, not gradients. SAM perturbations encode sharpness-sensitive directions that guide the optimizer, making their spectral structure directly tied to geometry rather than learning signals.
(2) Our goal is not compression but drift suppression: by removing low-frequency components, which we  empirically identify as the main carriers of inter-client heterogeneity,  we obtain more consistent client updates and promote convergence to flatter, more robust minima.

Let's consider a single local update step for a client $k$ at communication round $t \in [1,T]$ and local iteration $e \in [1,E]$. For simplicity, we omit the round and local iteration indices. Now, we first need to compute the standard SAM perturbation $\delta^k$ for client $k$ like FedSAM, denoted as 
    \begin{equation}\label{compute_delta}
    \centering
    \delta^k = \rho \cdot \frac{\nabla \mathcal{L}_k(w^k)}{\|\nabla\mathcal{L}_k(w^k)\|_2}
    \end{equation}

where $\delta^k$ has the same size with model parameter $w^k$ and includes all layer-wise perturbations $\delta^k_{1:L}$ and $\delta^k_l$ is the perturbation of parameter $w^k_l$ at layer $l$. The core idea of FedFFT is to {selectively discard the discordant low-frequency components} of the SAM perturbation while preserving the high-frequency components that capture more consistent aspects of the sharpness landscape. This filtering is performed on a per-layer basis to respect the model's architectural integrity. For any given weight layer $l$ with parameters $w_l^k$, the FedFFT procedure is as follows:

\label{ref:method}
\begin{enumerate}
    
    \item \textbf{Transform to Frequency Domain}: Instead of directly applying this perturbation, we transform it into the frequency domain. We flatten the perturbation tensor $\delta_l^k$ into a vector $\mathbf{v}_l^k$ and apply the Real-valued Fast Fourier Transform (rFFT):
    \begin{equation}
        \hat{\mathbf{v}}_l^k = \text{rFFT}(\mathbf{v}_l^k),
    \end{equation}
    where $\hat{\mathbf{v}}_l^k$ is the frequency-domain representation of the perturbation.

    \item \textbf{Apply High-Pass Filter}: Next, we apply a high-pass filter operator, $\mathcal{H}_r(\cdot)$, which zeroes out the lowest $r$ fraction of frequency coefficients. Given a truncation ratio $r \in [0, 1)$, the filtering operation is defined as:
    \begin{equation}
        [\mathcal{H}_r(\hat{\mathbf{v}}_l^k)]_m = 
        \begin{cases} 
            0, & \text{if } m < \lfloor r \cdot \text{len}(\hat{\mathbf{v}}_l^k) \rfloor, \\ 
            [\hat{\mathbf{v}}_l^k]_m, & \text{otherwise} ,
        \end{cases}
    \end{equation}
    where $m$ indexes the frequency coefficients in ascending order. This step is the crux of our method, as it explicitly removes the client-specific biases encoded in the low-frequency domain.

    \item \textbf{Reconstruct Filtered Perturbation}: We then transform the filtered vector back into the parameter domain using the inverse rFFT (iRFFT) and reshape it to its original tensor shape to obtain the refined perturbation $\tilde{\delta}_l$:
    \begin{equation}
        \tilde{\delta}_l^k = \text{reshape}(\text{iRFFT}(\mathcal{H}_r(\hat{\mathbf{v}}_l^k))).
    \end{equation}

\end{enumerate}

For simplicity, for any given layer's perturbation tensor $\delta_l^k$ at client $k$, we summarize the three steps above as a filtering operation:
\begin{equation}\label{eq:filter_operator}
    \tilde{\delta}_l^k = \text{Filter}(\delta_l^k,r).
\end{equation}
Finally, the local model in client $k$ is updated using this filtered perturbation, following the standard SAM procedure:
    \begin{equation}
        w^k \leftarrow w^k - \eta \cdot \nabla \mathcal{L}_k(w^k + \tilde{\delta}^k),
    \end{equation}
    where $\tilde{\delta}^k$ is the collection of all layer-wise filtered perturbations $\tilde{\delta}_l^k$.

By replacing the standard SAM perturbation with our filtered version, FedFFT forces the local optimizer to ignore the most heterogeneous directions of sharpness, thereby promoting greater consistency among client updates and facilitating convergence to a more robust global minimum. As shown in, we provide the workflow of FedAvg, FedSAM and our FedFFT in Algorithm  \ref{alg:fedfft}. 


\subsection{Complexity Analysis}
  \textbf{Standard SAM.}
    Let $d$ denote the number of parameters and $T_{\text{fw/bw}}$ denote the cost of a forward + backward pass.
    SAM requires two such passes per iteration: (i) computing gradients and applying the perturbation, and (ii) recomputing gradients at the perturbed parameters.
    The perturbation itself is linear in $d$.
    Thus, the total complexity is
    \[
    T_{\text{SAM}} = 2 T_{\text{fw/bw}} + O(d),
    \]
    where the $O(d)$ term is negligible compared with forward/backward computation.
    
    \textbf{Ours.}
    Our method modifies only the perturbation step. For each parameter tensor, we perform:
    (i) flattening the perturbation vector,  
    (ii) forward FFT $O(d \log d)$,  
    (iii) frequency-domain masking $O(d)$,  
    (iv) inverse FFT $O(d \log d)$.  
    This does not modify the forward/backward computation.  
    Thus,
    \[
    T_{\text{FFT}} = 2 T_{\text{fw/bw}} + O(d \log d).
    \]
    
    \textbf{Comparison.}
    Both SAM and FFT share the same dominant term $2 T_{\text{fw/bw}}$.  
    The additional $O(d \log d)$ FFT overhead is small in practice because  
    (i) $T_{\text{fw/bw}} \gg d\log d$ for modern deep networks, and  
    (ii) FFT operations are highly optimized on GPUs.  
    Therefore, FFT-SAM maintains the computational structure of SAM while adding only a minor cost.

\begin{algorithm}[t]
    \centering
    \renewcommand{\algorithmicrequire}{\textbf{Input:}}
    \renewcommand{\algorithmicensure}{\textbf{Output:}}
    \caption{FedAvg, FedSAM, and Our FedFFT}
    \label{alg:fedfft}
    \begin{algorithmic}[1]
        \REQUIRE Communication rounds \( T \), local epochs \( E \), perturbation radius \( \rho \), local learning rate \( \eta \), frequency truncation ratio \( r \)
        \ENSURE Global model \( w_{g}^{T} \)
        \STATE Initialize global model \( w_{g}^{0} \)
        \FOR{$t = 0$ to $T-1$}
            \STATE Randomly select active client set \( S_t \)
            \FOR{each client \( k \in S_t \) \textbf{in parallel}}
                \STATE \( w^{k,t,0} \leftarrow w_{g}^{t} \)
                \FOR{$e = 0$ to $E-1$}
                    \STATE \textit{perturbation stage}
                    \STATE \textcolor{gray}{\textbf{FedAvg:} \( \delta^{k,t,e} = 0 \)}
                    \STATE \textcolor{gray}{\textbf{FedSAM:} \( \delta^{k,t,e} = \rho \cdot \frac{ \nabla\mathcal{L}_k(w^{k,t,e})}{\|\nabla\mathcal{L}_k(w^{k,t,e})\|_2} \)}
                    \STATE \textcolor{blue}{\textbf{FedFFT:} \( \delta_{1:L}^{k,t,e} = \text{Filter}(\delta_{1:L}^{k,t,e}, r) \)}
                    \STATE \( w^{k,t,e+1} = w^{k,t,e} - \eta \nabla \mathcal{L}_k(w^{k,t,e} + \delta^{k,t,e}) \)
                \ENDFOR
                \STATE Send local model \( w^{k,t,E} \) to server
            \ENDFOR
            \STATE \( w_{g}^{t+1} \leftarrow \frac{1}{|S_t|} \sum_{k \in S_t} w^{k,t,E} \)
        \ENDFOR
        \STATE Return \( w_{g}^{T} \)
    \end{algorithmic}
\end{algorithm}

\subsection{SAM in FL}

To comprehensively understand the differences among existing SAM-based federated learning optimization methods, we have systematically summarized representative algorithms in Table ~\ref{tab:sam_fl_summary}. A comparison is made from the perspectives of base algorithms (FedAvg, Scaffold, FedDyn), optimization objectives (local or global sharpness), and perturbation construction methods, clearly revealing the differences in their design philosophies. This summary also further demonstrates the uniqueness of FedFFT in perturbation frequency domain filtering, complementing existing methods.
\begin{table}[t]
\centering
\caption{Summary of representative SAM-based federated learning algorithms addressing data heterogeneity.}
\label{tab:sam_fl_summary}
\renewcommand{\arraystretch}{1.2}
\small 
\begin{tabular}{p{2.5cm} p{2.3cm} p{3.8cm} p{4.8cm}}
\toprule
\textbf{Research Work} & \textbf{Base Algorithm} & \textbf{Minimizing Target} & \textbf{Perturbation Formulation} \\
\midrule

FedSAM (\cite{qusam,sam}) 
& FedAvg 
& Local sharpness 
& $\rho \cdot \dfrac{\nabla \mathcal{L}_k}{\|\nabla \mathcal{L}_k\|}$ \\

MoFedSAM (\cite{qusam}) 
& FedAvg + Momentum 
& Local sharpness
& $\rho \cdot \dfrac{\nabla \mathcal{L}_k}{\|\nabla \mathcal{L}_k\|}$ \\

FedGAMMA (\cite{fedgamma})
& Scaffold
& Local sharpness
& $\rho \cdot \dfrac{\nabla \mathcal{L}_k}{\|\nabla \mathcal{L}_k\|}$ \\

FedSMOO (\cite{fedsmoo})
& FedDyn
& Local sharpness (corrected)
& $\rho \cdot \dfrac{\nabla \mathcal{L}_k - \mu_i - s}{\|\nabla \mathcal{L}_k - \mu_i - s\|}$ \\

FedLESAM (\cite{fedlesam})
& FedAvg / Scaffold / FedDyn
& Global sharpness
& $\rho \cdot \dfrac{w^{k}_{old} - w_g^t}{\|w^{k}_{old} - w_g^t\|}$ \\

FedGloSS (\cite{fedgloss})
& FedDyn-like
& Global sharpness via pseudo-gradient
& $\rho \cdot \dfrac{\Delta^{t-1}_w}{\|\Delta^{t-1}_w\|}$ \\

FedFFT (ours)
& FedAvg / Scaffold / FedDyn
& Local sharpness with spectral filtering
& $\rho \cdot \text{Filter}\!\left(\dfrac{\nabla \mathcal{L}_k}{\|\nabla \mathcal{L}_k\|}\right)$ \\
\bottomrule
\end{tabular}
\end{table}

\section{Experiments}
\label{Experiment}

\subsection{Experiments Setup}

\subsubsection{Datasets} We evaluate our methods on three widely used image classification benchmarks: CIFAR-10 \cite{cifar}, CIFAR-100 \cite{cifar}, and Tiny-ImageNet \cite{tinyimagenet}. These datasets vary in complexity, number of classes, and dataset size, providing a comprehensive evaluation of the robustness and generalization of federated learning algorithms. Specifically, CIFAR-10 contains 60,000 color images of size 32×32 pixels across 10 classes, with 50,000 training images and 10,000 test images. CIFAR-100 has the same total number of images but includes 100 classes, with 500 training images and 100 test images per class, making it a more challenging benchmark due to the increased number of classes and finer-grained categories. Tiny-ImageNet is a subset of the ImageNet dataset, consisting of 200 classes, each with 500 training images, 50 validation images, and 50 test images, with images resized to 64×64 pixels. Its larger number of classes and higher-resolution images present a significantly more challenging scenario for federated learning.

To simulate realistic data heterogeneity across clients, we adopt a Dirichlet distribution-based partitioning strategy. Specifically, for each dataset, we sample class distributions for each client from a Dirichlet distribution parameterized by $\alpha \in {0.1, 0.6}$. A smaller $\alpha$ value corresponds to more non-IID distributions, meaning that individual clients receive highly skewed subsets of classes, reflecting the imbalanced and biased data often observed in real-world federated scenarios. Conversely, a larger $\alpha$ value leads to more IID-like distributions, where each client’s local dataset more closely approximates the overall class distribution. To illustrate the severity of data heterogeneity under the highly non-IID setting, Figure~\ref{fig:data_distribution_alpha} visualizes the client-wise class distributions of CIFAR-10, CIFAR-100, and Tiny-ImageNet under Dirichlet partitions with $\alpha = 0.1$ and $\alpha = 0.6$. Each heatmap illustrates the proportion of samples per class for each client, where the horizontal axis corresponds to client indices and the vertical axis represents class IDs. As expected, the $\alpha = 0.1$ setting exhibits highly skewed and imbalanced data allocations: certain clients show dark, concentrated regions indicating dominance of only a few classes, while other classes may be nearly absent. This reflects a strongly non-IID scenario commonly observed in real-world FL applications. In contrast, when $\alpha = 0.6$, the heatmaps become noticeably smoother with reduced color variations, indicating that clients receive more balanced class distributions and the overall data heterogeneity is significantly alleviated. These two complementary visualizations allow us to systematically assess the behavior and robustness of different federated learning algorithms across varying levels of client data heterogeneity.

\vspace{1em} 
\noindent\begin{minipage}{\textwidth}
    \centering
    
    \begin{minipage}{0.31\textwidth}
        \centering
        \includegraphics[width=\linewidth]{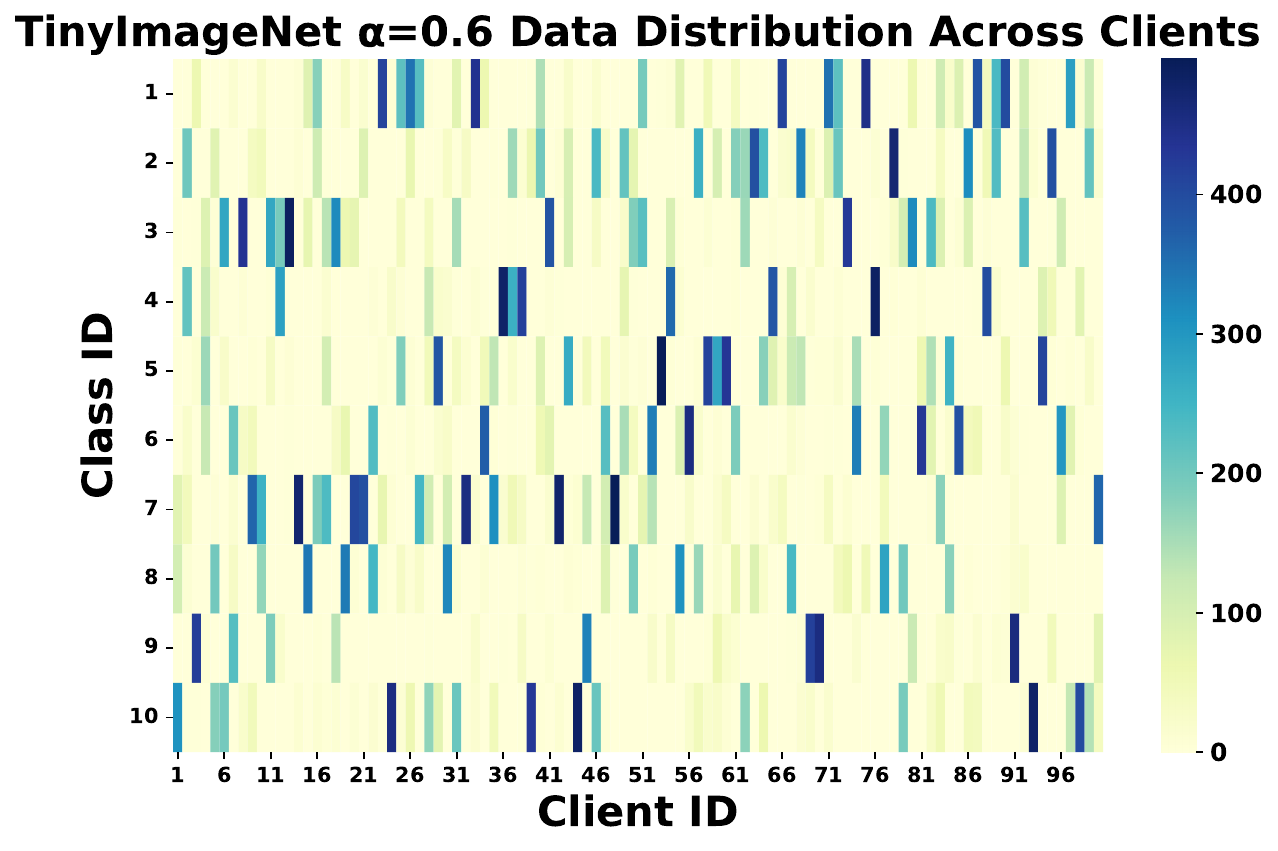}
        \centerline{\small (a) CIFAR-10, $\alpha=0.1$}
        \label{fig:cifar10_0.1}
    \end{minipage}\hfill
    \begin{minipage}{0.31\textwidth}
        \centering
        \includegraphics[width=\linewidth]{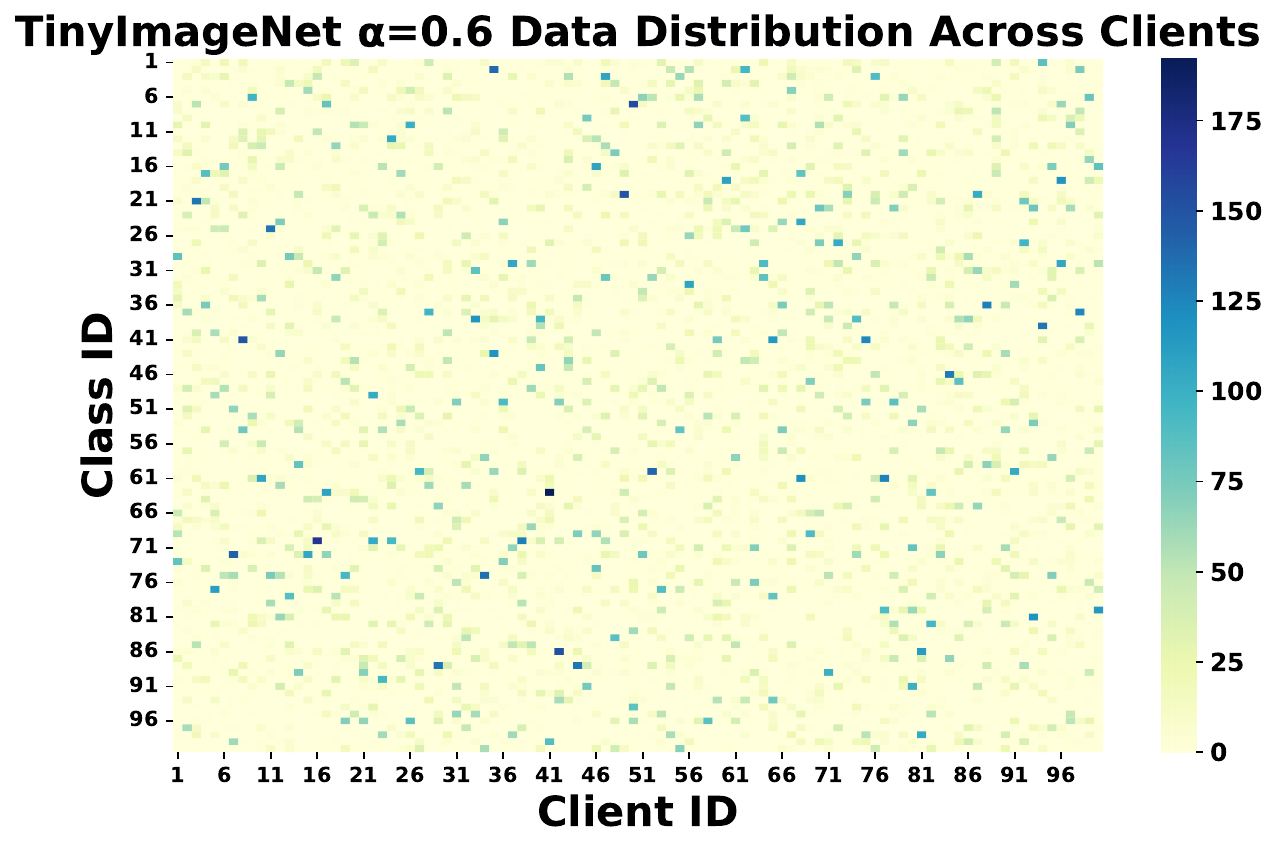}
        \centerline{\small (b) CIFAR-100, $\alpha=0.1$}
        \label{fig:cifar100_0.1}
    \end{minipage}\hfill
    \begin{minipage}{0.31\textwidth}
        \centering
        \includegraphics[width=\linewidth]{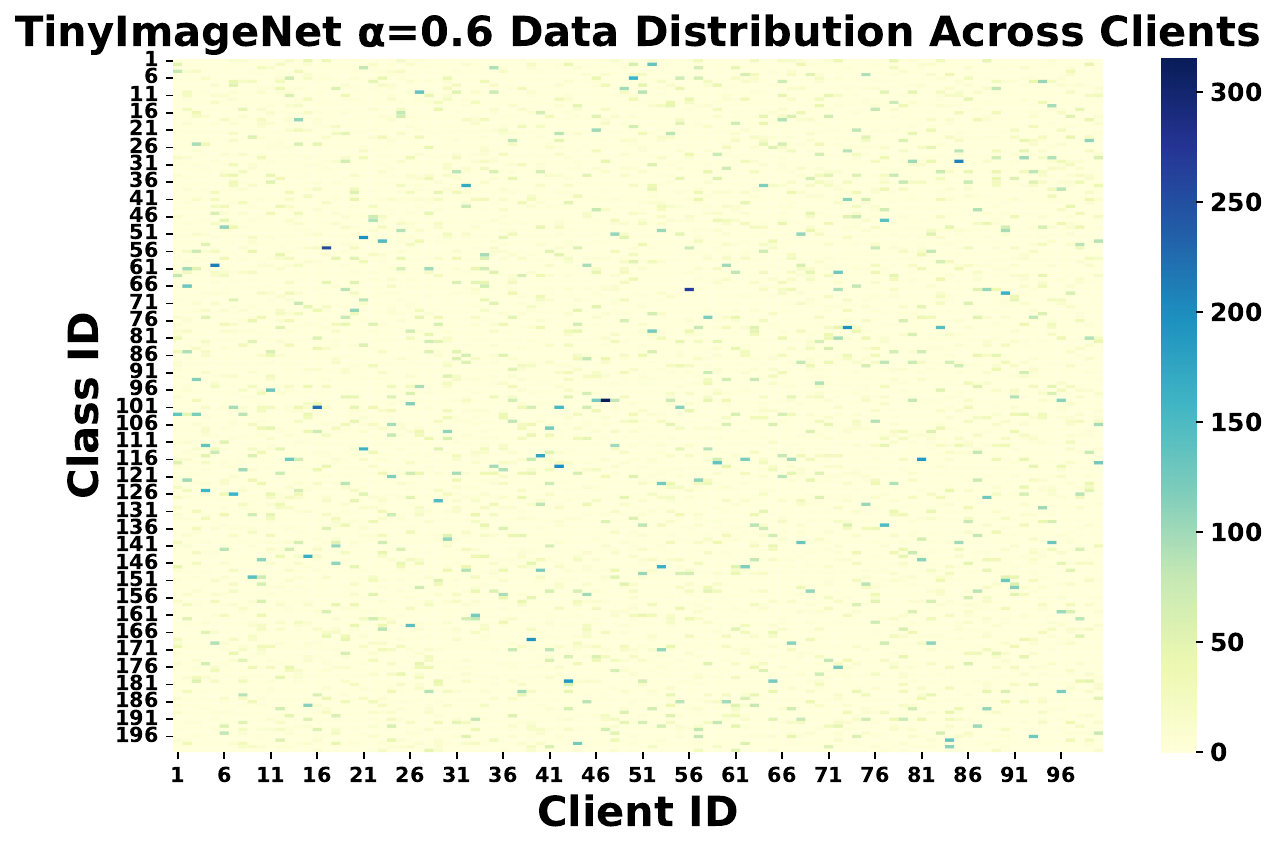}
        \centerline{\small (c) Tiny ImageNet, $\alpha=0.1$}
        \label{fig:tinyimagenet_0.1}
    \end{minipage}
    
    \vspace{15pt} 

    \begin{minipage}{0.31\textwidth}
        \centering
        \includegraphics[width=\linewidth]{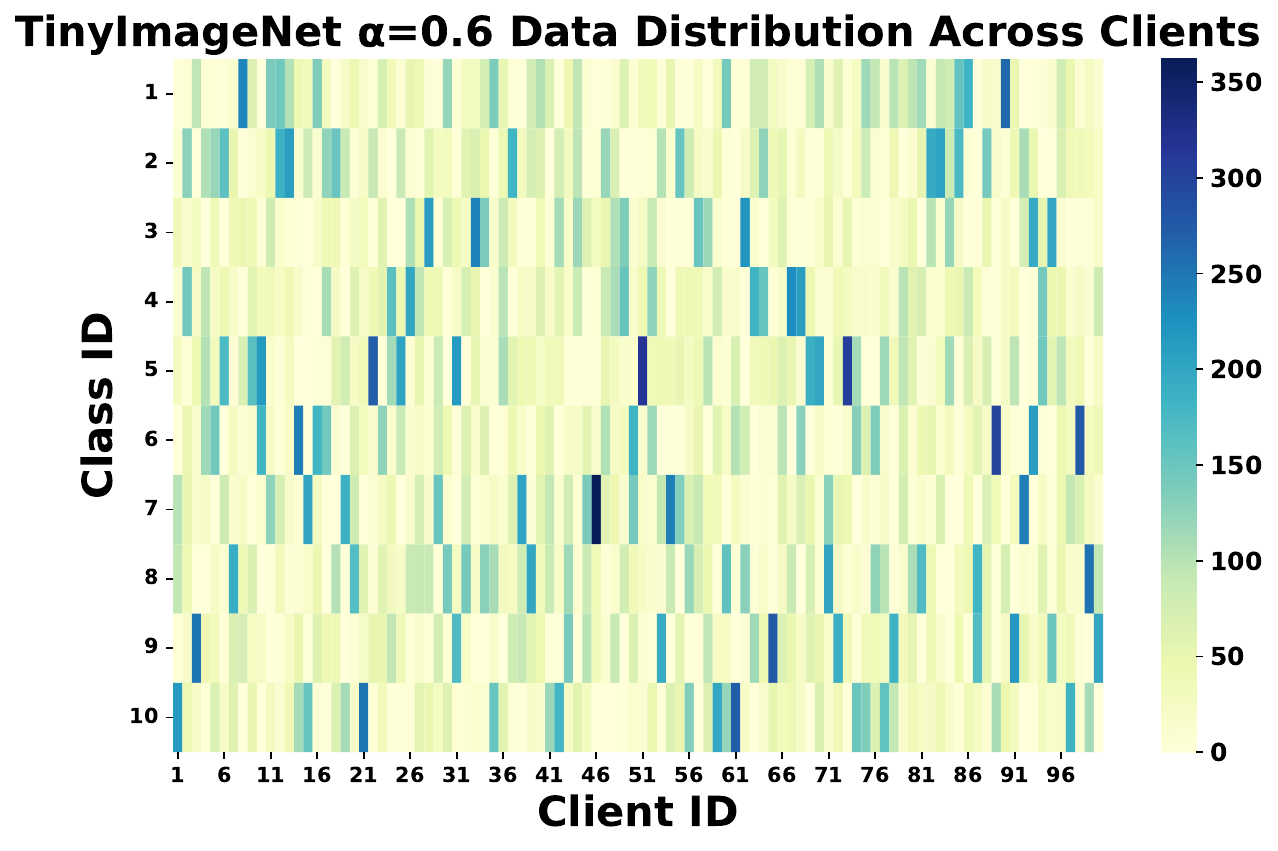}
        \centerline{\small (d) CIFAR-10, $\alpha=0.6$}
        \label{fig:cifar10_0.6}
    \end{minipage}\hfill
    \begin{minipage}{0.31\textwidth}
        \centering
        \includegraphics[width=\linewidth]{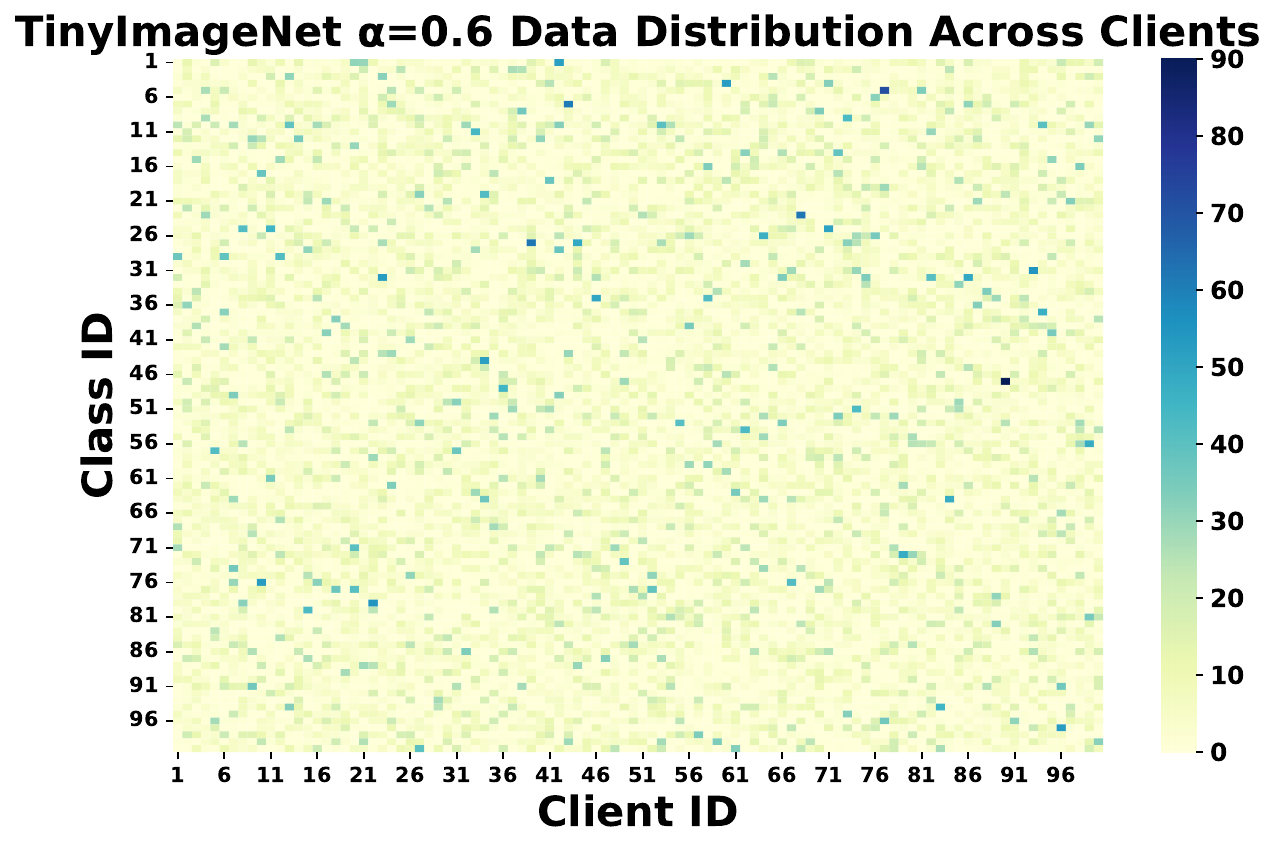}
        \centerline{\small (e) CIFAR-100, $\alpha=0.6$}
        \label{fig:cifar100_0.6}
    \end{minipage}\hfill
    \begin{minipage}{0.31\textwidth}
        \centering
        \includegraphics[width=\linewidth]{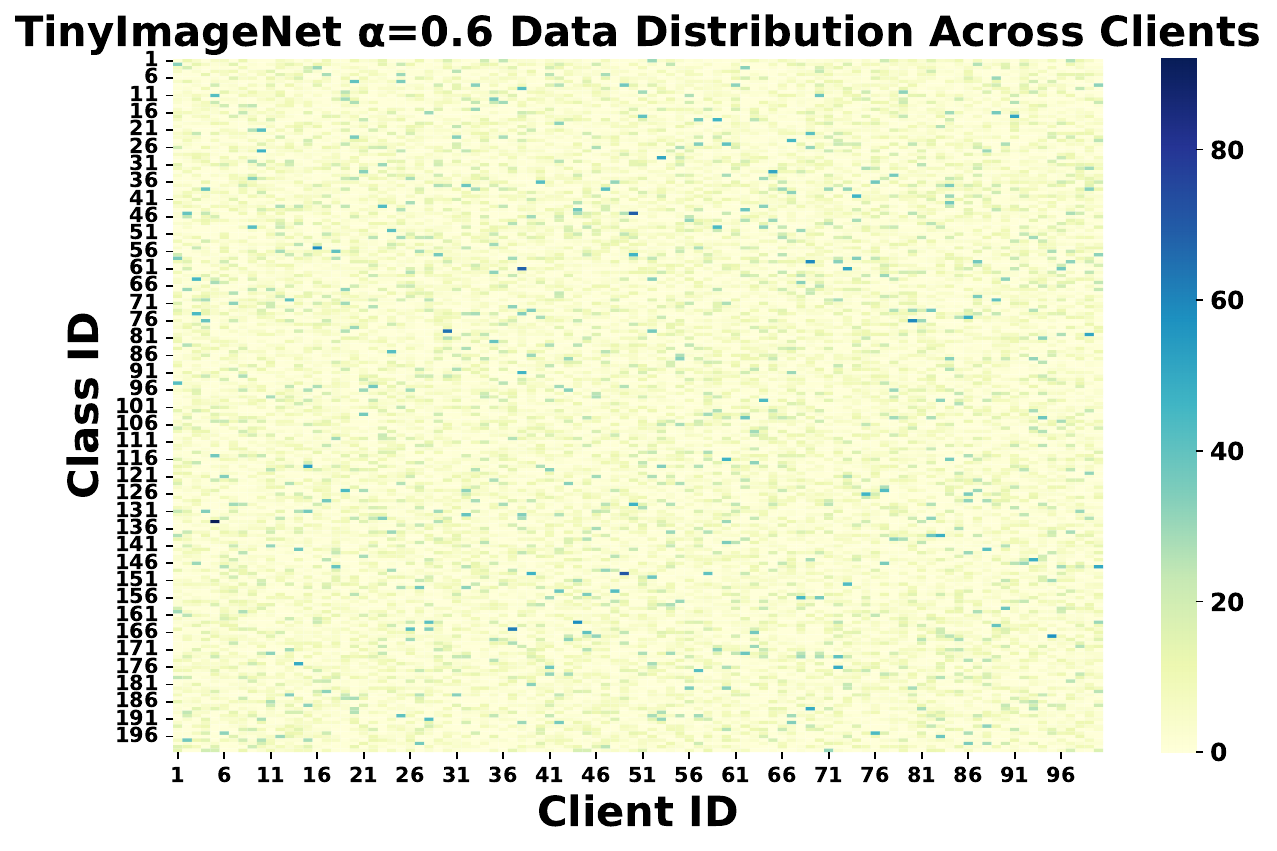}
        \centerline{\small (f) Tiny ImageNet, $\alpha=0.6$}
        \label{fig:tinyimagenet_0.6}
    \end{minipage}

    \captionof{figure}{Client data distributions across CIFAR-10, CIFAR-100, and Tiny-ImageNet under Dirichlet partitions $\alpha=0.1$ and $\alpha=0.6$.}
    \label{fig:data_distribution_alpha}
\end{minipage}
\vspace{1em} 

\subsubsection{Baselines}

We benchmark FedFFT against a comprehensive suite of baselines in three main categories: (i) foundational FL algorithms, including FedAvg~\cite{fedavg}, SCAFFOLD~\cite{scaffold}, and FedDyn~\cite{feddyn}; (ii) the direct application of SAM in FL, namely FedSAM and its momentum-enhanced variant MoFedSAM~\cite{caldarola2022improving,qusam}; and (iii) advanced FL-SAM variants that aim to improve consistency, such as FedGAMMA~\cite{fedgamma}, FedSMOO~\cite{fedsmoo}, FedLESAM~\cite{fedlesam}, and FedGloSS~\cite{fedgloss}. To ensure fair comparisons, we integrate our approach with these foundational algorithms following FedLESAM, yielding FedFFT (FedAvg-based), FedFFT-S (SCAFFOLD-based), and FedFFT-D (FedDyn-based). We summarize the characteristics of SAM in FL methods; please refer to the Table ~\ref{tab:sam_fl_summary} for details.

\subsubsection{Backbone}

For the main experiments, the backbone network is ResNet-18 equipped with Group Normalization (GN) layers instead of Batch Normalization (BN) to improve training stability in federated settings. GN alleviates the issues caused by small and heterogeneous local batch sizes, making it more suitable for cross-client training.

To further evaluate the generality and robustness of our approach across different model architectures, we additionally conduct cross-model experiments using three representative backbones: ResNet-20, a CIFAR-specific variant where all BN layers are replaced with GN to enhance stability under non-IID data; DenseNet-121, implemented following the standard DenseNet design with GN applied after the final dense block; and Vision Transformer (ViT), where we adopt the \texttt{vit\_tiny\_patch16\_224} model from the \texttt{timm} library to evaluate our method on transformer-based architectures.

These diverse architectures allow us to assess whether the proposed method generalizes effectively beyond a single CNN backbone.

\subsubsection{Implementation Details}
For our experiments on CIFAR-10 \cite{cifar}, CIFAR-100 \cite{cifar}, and Tiny-ImageNet \cite{tinyimagenet}, we adopt training configurations consistent with FedSMOO \cite{fedsmoo} and FedLESAM \cite{fedlesam} to ensure fair comparison. The backbone network is ResNet-18 equipped with Group Normalization (GN) layers instead of Batch Normalization (BN) to improve training stability in federated settings. Optimization is performed using stochastic gradient descent (SGD). The total number of communication rounds is set to 800 for CIFAR-10 and CIFAR-100, and 300 for Tiny-ImageNet. The initial local learning rate is $\eta = 0.1$. Unless otherwise specified, the learning rate decays exponentially by a factor of $0.998\times$ per round; however, FedDyn, FedSMOO, FedLESAM-D, and FedFFT-D use a slower decay rate of $0.9995\times$ for the proxy term. The local training settings for each dataset are as follows: for CIFAR-10, we use a batch size of 50 and 5 local epochs; for CIFAR-100, the batch size is 20 with 2 local epochs; for Tiny-ImageNet, we follow the same configuration as CIFAR-10. These choices balance computational efficiency and convergence in federated settings.

\vspace{1em} 
\noindent\begin{minipage}{\textwidth}
    \centering
    
    \begin{minipage}{0.31\textwidth}
        \centering
        \includegraphics[width=\linewidth]{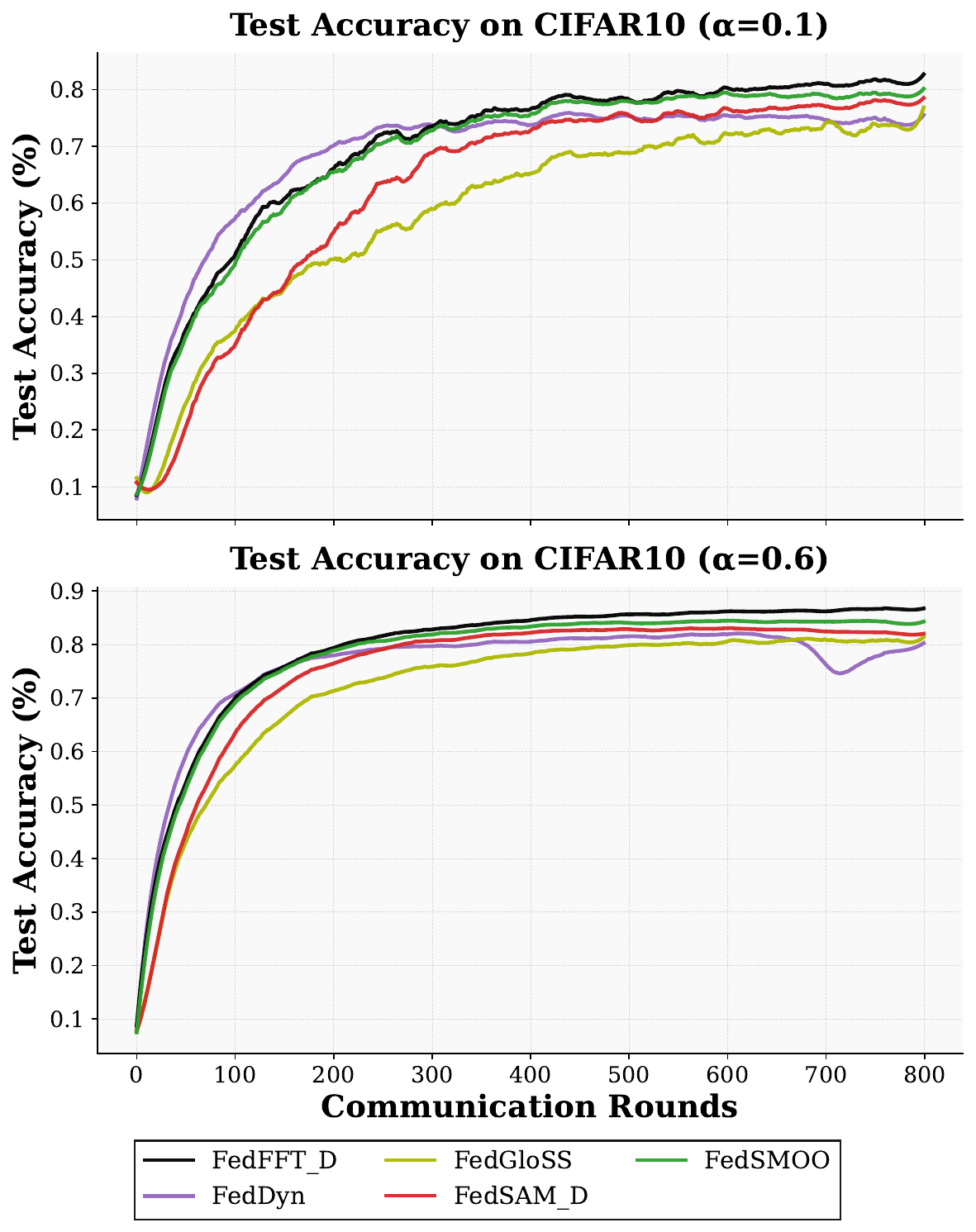}
        \centerline{\small (a) CIFAR-10}
    \end{minipage}\hfill
    \begin{minipage}{0.31\textwidth}
        \centering
        \includegraphics[width=\linewidth]{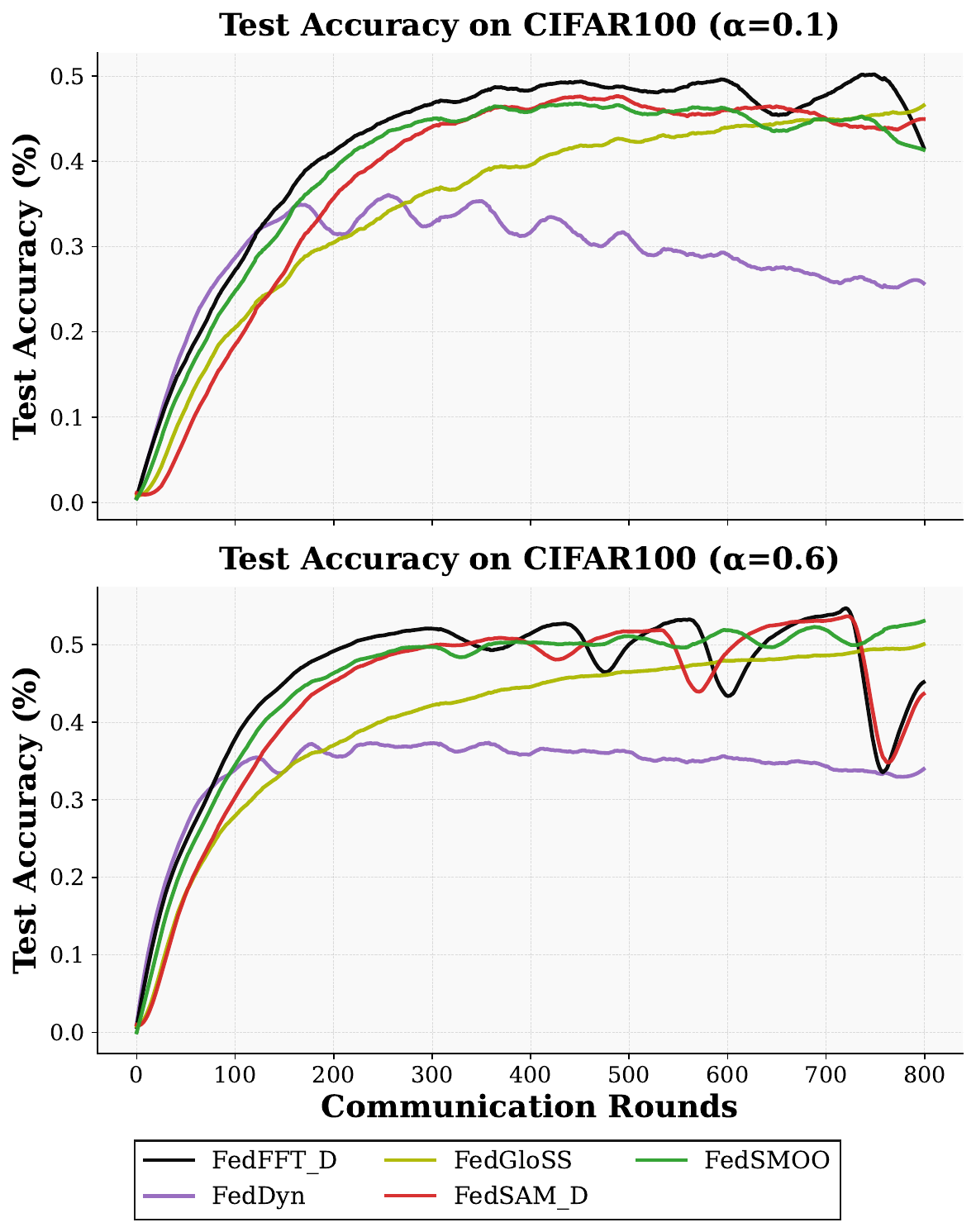}
        \centerline{\small (b) CIFAR-100}
    \end{minipage}\hfill
    \begin{minipage}{0.31\textwidth}
        \centering
        \includegraphics[width=\linewidth]{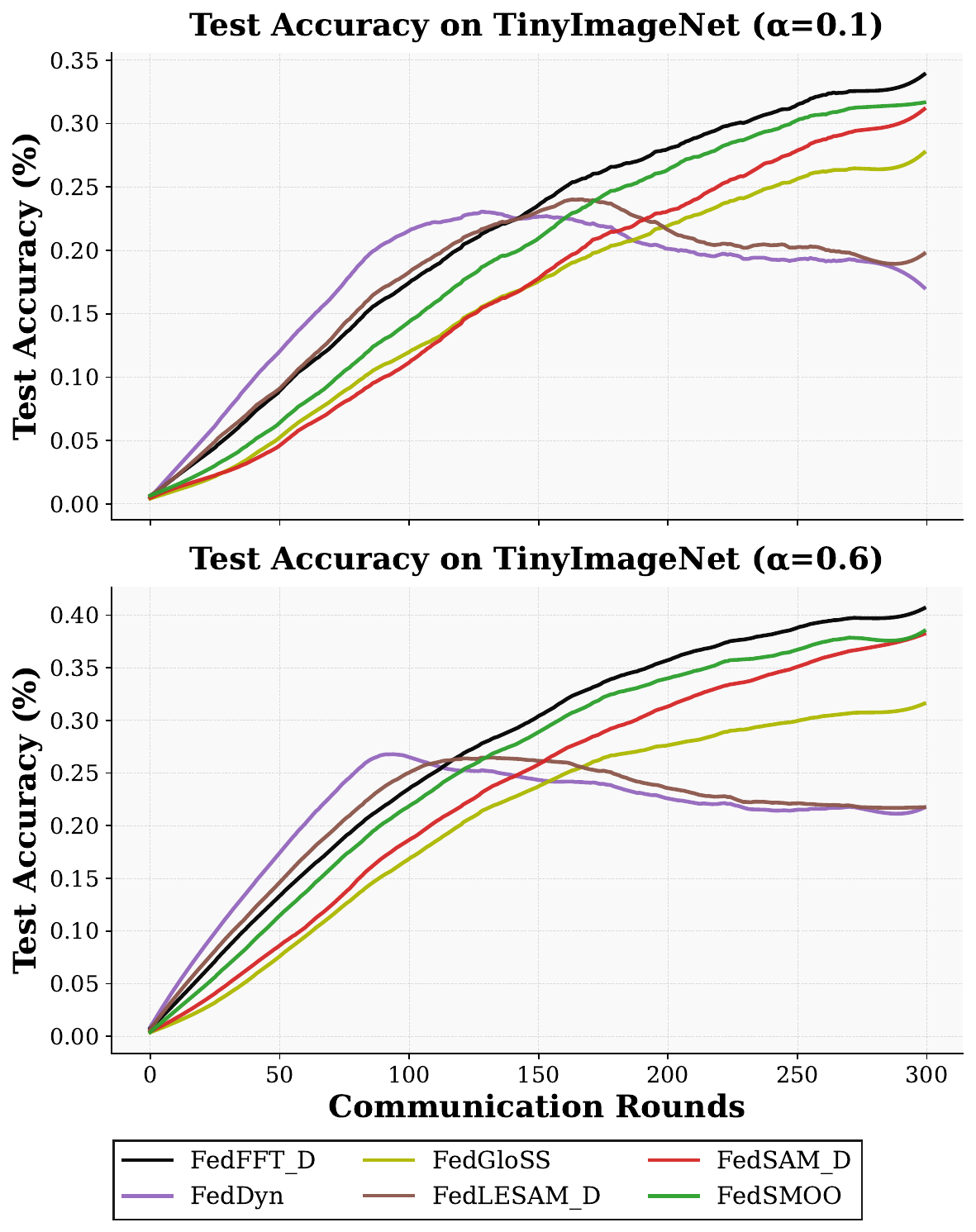}
        \centerline{\small (c) Tiny-ImageNet}
    \end{minipage}

    \captionof{figure}{\small Learning curves on three benchmarks under Dirichlet partitionings with concentration parameters 0.1 and 0.6, using a ResNet-18 model.}
    \label{fig:learning_curve}
\end{minipage}
\vspace{1em} 

\subsection{Main Results}

\textbf{Comparison with State-of-the-Art Baselines.} Table~\ref{tab:table-main} presents the results on CIFAR-10, CIFAR-100, and Tiny-ImageNet with ResNet-18. Overall, FedFFT delivers consistent improvements under both SCAFFOLD and FedDyn frameworks, indicating that frequency-domain perturbation modeling complements existing federated optimization paradigms. On CIFAR-10, FedFFT-D achieves the best performance under both Dirichlet partitions, demonstrating its effectiveness in standard federated settings. For CIFAR-100, while FedFFT-D is slightly outperformed by FedLESAM-S under $\alpha=0.6$, it regains superiority in the more heterogeneous $\alpha=0.1$ case, suggesting that suppressing low-frequency perturbations is particularly beneficial when client data distributions are highly skewed. Moreover, on Tiny-ImageNet, FedFFT-D clearly surpasses all baselines, highlighting its robustness and scalability to larger and more challenging tasks.

\begin{table}[H]  
\centering
\small  
\caption{Test accuracy comparison (\%) of different methods on CIFAR-10, CIFAR-100 and Tiny-ImageNet, with ResNet-18. ``-S'' and ``-D'' denote using SCAFFOLD and FedDyn as base algorithms. All results on Tiny-ImageNet are reproduced by us. Results marked with {\dag} are reproduced by us for CIFAR-10 and CIFAR-100. Others are from \cite{fedsmoo,fedlesam}.}
\label{tab:table-main}
\begin{tabular}{@{}lcccccc@{}}
\toprule
\multirow{2}{*}{Method} & \multicolumn{2}{c}{CIFAR-10} & \multicolumn{2}{c}{CIFAR-100} & \multicolumn{2}{c}{Tiny-ImageNet} \\
\cmidrule(lr){2-3} \cmidrule(lr){4-5} \cmidrule(l){6-7}
 & $\alpha=0.6$ & $\alpha=0.1$ & $\alpha=0.6$ & $\alpha=0.1$ & $\alpha=0.6$ & $\alpha=0.1$ \\
\midrule
FedAvg & 79.52 & 76.00 & 46.35 & 42.64 & 28.31 & 27.48 \\
FedSAM{\dag} & 81.91 & 74.92 & 48.08 & 45.53 & 33.16 & 29.46 \\
MoFedSAM & 84.13 & 78.71 & 54.38 & 44.85 & 33.50 & 29.77 \\
FedLESAM & 81.04 & 76.93 & 47.92 & 44.48 & 27.91 & 26.91 \\
Our FedFFT & 83.02 & 77.53 & 48.59  & 46.83 & 33.58 & 30.43 \\
\midrule
SCAFFOLD &81.81 &78.57 & 51.98 & 44.41 & 35.34 & 32.11\\
FedSAM-S{\dag}  & 83.88 & 76.68&50.19&49.14&35.84&31.73\\
FedGamma-S & 82.64 & 78.95 & 53.41 & 46.39 & 36.85 & 30.09 \\
FedLESAM-S & 84.94 & 79.52 & \textbf{54.61} & 48.07 & 28.47 & 27.70 \\
Our FedFFT-S & 84.69 & 79.24 & 52.75 & \underline{49.85} & 36.15 & \underline{33.08} \\
\midrule
FedDyn & 83.22 & 78.08 & 50.82 & 42.50 & 28.01 & 24.19 \\
FedSAM-D{\dag}  & 82.29 & 79.11 & 53.70 & 46.28 &38.18&31.39\\
FedSMOO-D & \underline{84.55} & \underline{80.82} & 53.92 & 46.48 & \underline{38.71} & 32.45 \\
FedLESAM-D & 84.27 & 80.08 & 53.27 & 46.42 & 27.36 & 25.32 \\
FedGloSS-D{\dag} & 82.58 & 79.23 & 50.92 & 47.36 & 31.72 & 28.04 \\
Our FedFFT-D & \textbf{87.19} & \textbf{83.05} & \underline{54.46} & \textbf{50.90} & \textbf{40.85} & \textbf{34.46} \\
\bottomrule
\end{tabular}
\end{table}

Beyond final accuracy, the convergence curves in Figure~\ref{fig:learning_curve} further underline the advantage of our method. Across all three datasets—CIFAR-10, CIFAR-100, and Tiny-ImageNet—FedFFT consistently converges faster than competing approaches. This accelerated convergence indicates that by harmonizing client updates at the spectral level, FedFFT navigates the optimization landscape more efficiently, reducing the number of communication rounds required to reconcile heterogeneous updates and stabilizing learning dynamics even under highly non-IID partitions.

\textbf{Generalization Across Diverse Model Architectures.} 
To verify that the efficacy of FedFFT is not confined to a specific model class, we evaluate its performance across a diverse range of architectures, from lightweight ResNet-18, ResNet-20 to deeper DenseNet-121 and Vision Transformers (ViT). We perform on CIFAR-10 and CIFAR-100 with a moderately heterogeneous setting $\alpha=0.6$. As shown in Tab ~\ref{tab:arch_results}, FedFFT usually outperforms baselines across different architectures. Critically, the performance gains are substantial even on powerful models like ViT. This suggests that the problem FedFFT addresses—the inconsistency in the low-frequency spectrum of client perturbations—is a fundamental artifact of the federated optimization process itself, independent of a model's representation capacity. Simply using a larger model does not automatically resolve the geometric misalignment between clients. Our spectral filtering acts as a complementary and orthogonal improvement, harmonizing the local updates to allow these powerful architectures to converge more effectively. These results therefore underscore the broad applicability and scalability of FedFFT, establishing it as a model-agnostic enhancement for sharpness-aware federated learning.

\textbf{Visualization of the Global Loss Landscape.} To gain deeper insight into the optimization behavior, we also visualize the 2D loss landscapes of different algorithms on CIFAR-10 with ResNet-18 under the challenging $\alpha=0.1$ setting (Figure~\ref{fig:2dcompare}). We compare four representative methods: FedSAM, FedSMOO, FedGLOSS, and our FedFFT-D. The visualized ranges of these landscapes reveal clear differences in sharpness: FedSAM exhibits a steep and narrow basin (range approximately 0.6–8.6), while FedSMOO (0.3–5.7) and FedGLOSS (0.45–5.85) show moderately improved smoothness. In contrast, FedFFT-D displays the widest and flattest landscape (0.3–4.8), with a noticeably broader and smoother central region. This indicates that FedFFT-D not only finds flatter minima but also maintains stability across a larger neighborhood, reflecting stronger generalization and a reduced tendency to overfit client-specific noise. Such landscape characteristics align well with the improved accuracy and convergence speed observed earlier, providing geometric evidence that spectral-level update alignment leads to more stable federated optimization.

\textbf{Visualizing the Effect of Spectral Filtering.}
To qualitatively understand how FedFFT achieves greater consistency, we visualize the distributions of client perturbations, model features, and model parameters. The visualizations in Fig.~\ref{fig:pca_comparsion} reveal a clear causal chain. \textit{(1) More Cohesive Perturbations (Fig. \ref{fig:pca_p}):} The process begins with the perturbations themselves. The original SAM perturbations from different clients are widely scattered in the PCA space. After applying FedFFT's low-frequency filter, these perturbations become significantly more compact, confirming that our method successfully reduces inter-client discrepancy at its source. \textit{(2) Aligned Feature Representations (Fig.~\ref{fig:pca_feature}):} This improved consistency in perturbations directly translates to more aligned model behavior. \textcolor{black}{We observe that the average features extracted by the FedFFT-trained model are much more tightly clustered for each class compared to the scattered features from the FedSAM model.} \textit{(3) Consolidated Client Models (Fig. \ref{fig:pca_params}):} Ultimately, this leads to better convergence of the models themselves. The parameters of client models trained with FedFFT exhibit a much smaller variance and are clustered closer to a central point, indicating that spectral filtering effectively mitigates client drift and guides all clients toward a more unified and robust global solution.

\vspace{0.5em} 
\noindent\begin{minipage}{\textwidth}
    \centering
    
    \begin{minipage}{0.24\textwidth}
        \centering
        \includegraphics[width=\linewidth]{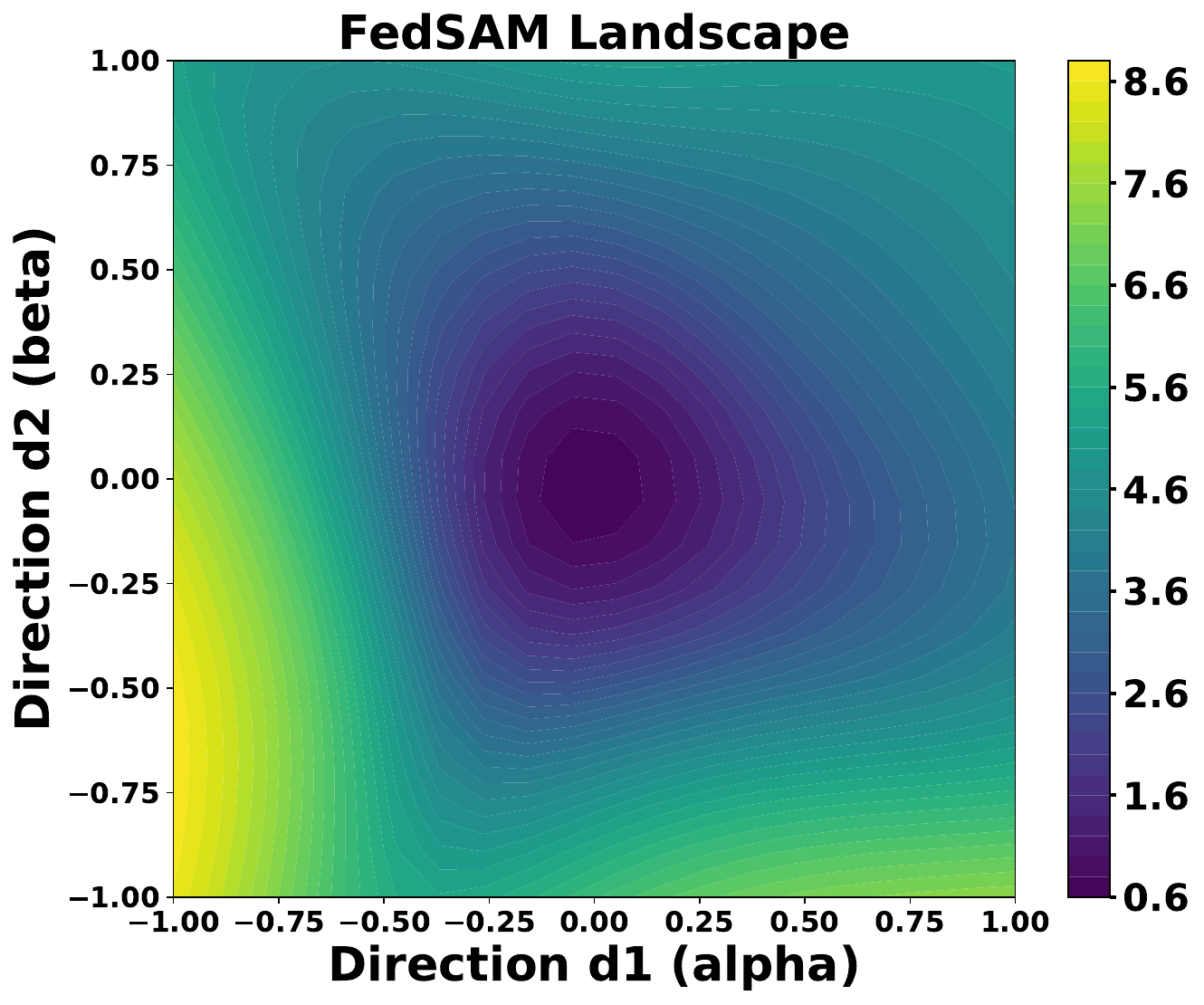}
        \centerline{\small (a) FedSAM}
    \end{minipage}\hfill
    \begin{minipage}{0.24\textwidth}
        \centering
        \includegraphics[width=\linewidth]{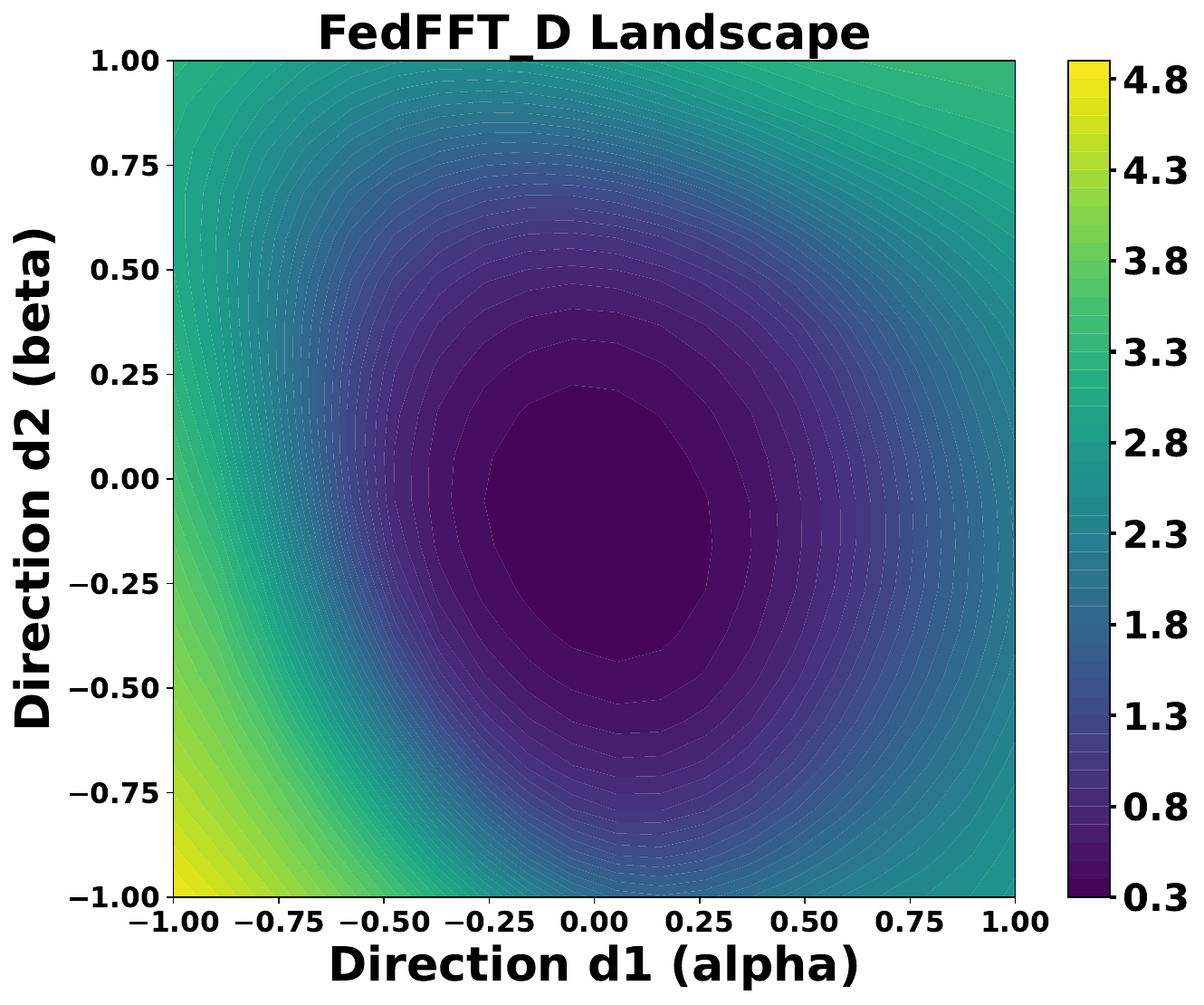}
        \centerline{\small (b) FedFFT\_D}
    \end{minipage}\hfill
    \begin{minipage}{0.24\textwidth}
        \centering
        \includegraphics[width=\linewidth]{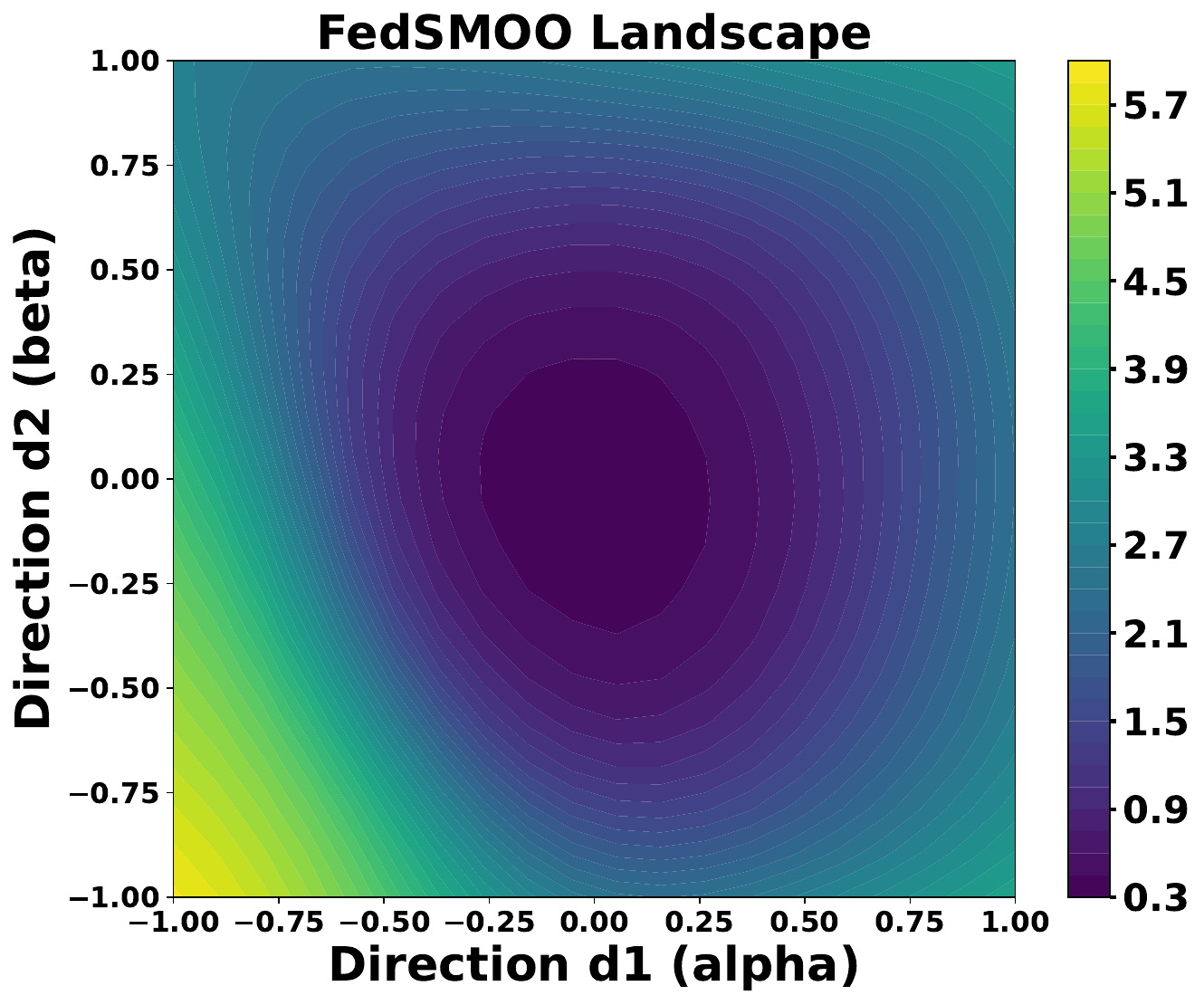}
        \centerline{\small (c) FedSMOO}
    \end{minipage}\hfill
    \begin{minipage}{0.24\textwidth}
        \centering
        \includegraphics[width=\linewidth]{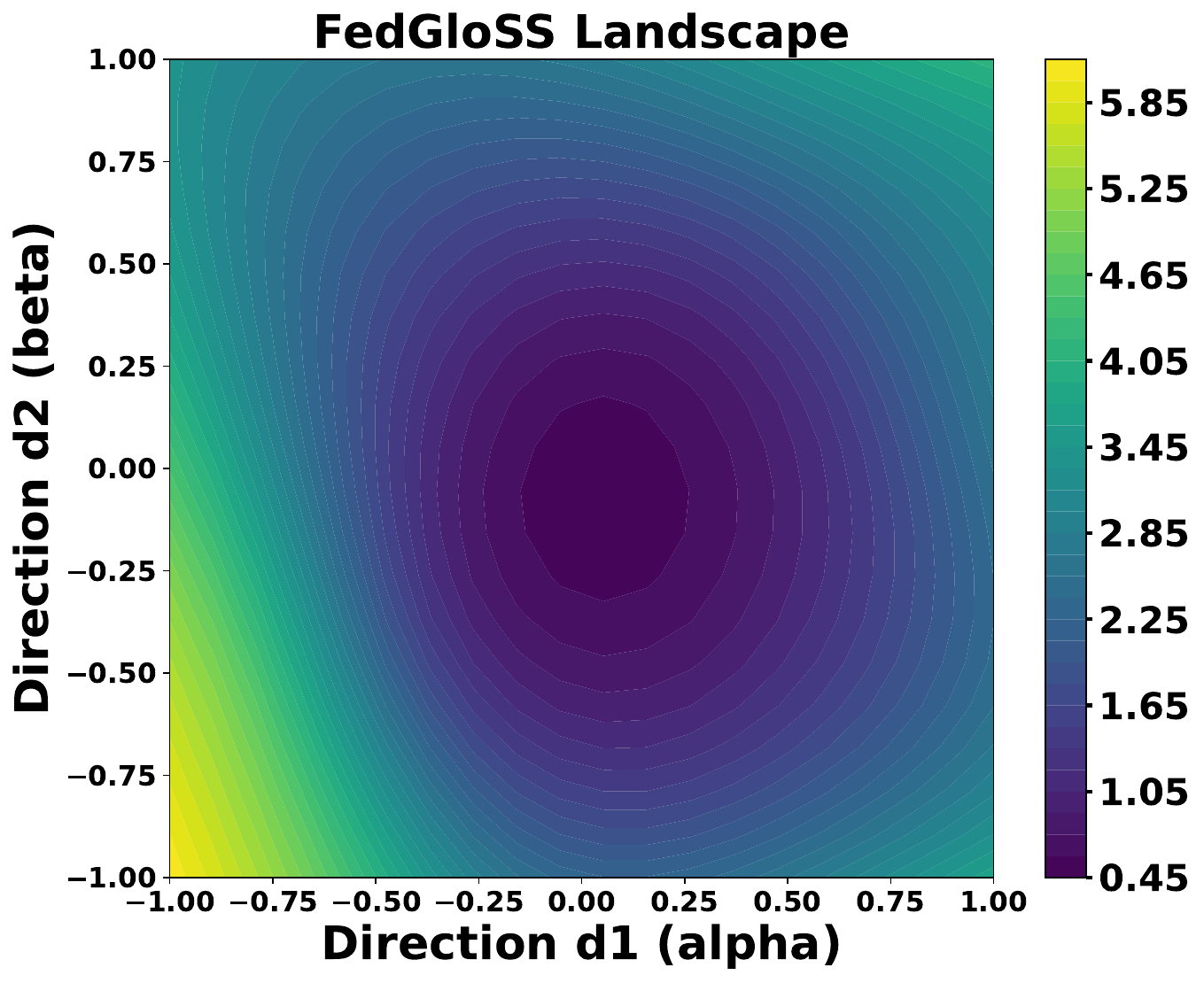}
        \centerline{\small (d) FedGloss}
    \end{minipage}

    \captionof{figure}{\small 2D loss landscape comparison across different algorithms on CIFAR-10 under Dirichlet partition with $\alpha = 0.1$, using ResNet-18 as the backbone model.}
    \label{fig:2dcompare}
\end{minipage}
\vspace{0.5em} 

\begin{table}[t] 
    \centering
    \footnotesize 
    \setlength{\tabcolsep}{4pt} 
    \renewcommand{\arraystretch}{1.0} 
    
    \caption{\small {Training wall-clock time to reach FedAvg accuracy on CIFAR-10 with Dirichlet $\alpha=0.1$ using ResNet-18. Experiments run on NVIDIA Tesla A40.}}
    \label{tab:efficiency}
    \begin{tabular}{lccc}
    \toprule
    \textbf{Method} & \textbf{Time/Round (s)} & \textbf{Rounds} & \textbf{Total Time (s)} \\ 
    \midrule
    FedAvg       & 11.73 & 800 & 9390.21 \\
    FedSAM       & 16.84 & 718 & 12093.24 \\
    FedSMOO      & 21.74 & 312 & 6785.71 \\
    FedGloSS     & 23.07 & 487 & 11235.70 \\
    \textbf{FedFFT-D (Ours)} & \textbf{21.63} & \textbf{302} & \textbf{6532.98} \\
    \bottomrule
    \end{tabular}
\end{table}

\textbf{Training Efficiency.}
Training efficiency is another crucial factor in federated learning (FL), as practical deployments often require methods to reach a target accuracy under limited computational and wall-clock budgets. In this study, we adopt FedAvg as the reference baseline and first run it for 800 rounds on CIFAR-10 under a Dirichlet–0.1 client distribution using a ResNet-18 backbone, recording the highest accuracy it achieves. For each competing method, we then evaluate its training efficiency in terms of (i) the number of communication rounds required to match this accuracy level, and (ii) the total wall-clock time consumed. The average per-round time (the “Times’’ column in Table~\ref{tab:efficiency}) is computed by dividing the total training time by the number of rounds. As shown in Table~\ref{tab:efficiency}, although FFT introduces a theoretically higher perturbation cost compared with the linear perturbation used in SAM, its practical overhead is negligible relative to the dominant forward/backward cost. The optimization benefits brought by our FFT-based perturbation filtering lead to significantly fewer required rounds. FedFFT-D reaches the target accuracy in only 302 rounds—representing a \textbf{62.25\% reduction} in communication rounds compared to FedAvg (800 rounds). Even relative to strong baselines such as FedSMOO (312 rounds) and FedGloSS (487 rounds), FedFFT-D still converges faster, requiring the fewest rounds among all methods. The accelerated convergence enabled by our FFT-based optimization leads to a substantial reduction in total wall-clock time. Overall, FedFFT-D achieves the fastest training time among all methods while reaching the same target accuracy as FedAvg. The suppression of low-frequency components in the perturbation appears to enable a more stable and direct optimization trajectory, thereby accelerating convergence under highly heterogeneous client distributions. 

\vspace{1em} 
\noindent\begin{minipage}{\textwidth}
    \centering
    
    \begin{minipage}{0.31\textwidth}
        \centering
        \includegraphics[width=\linewidth]{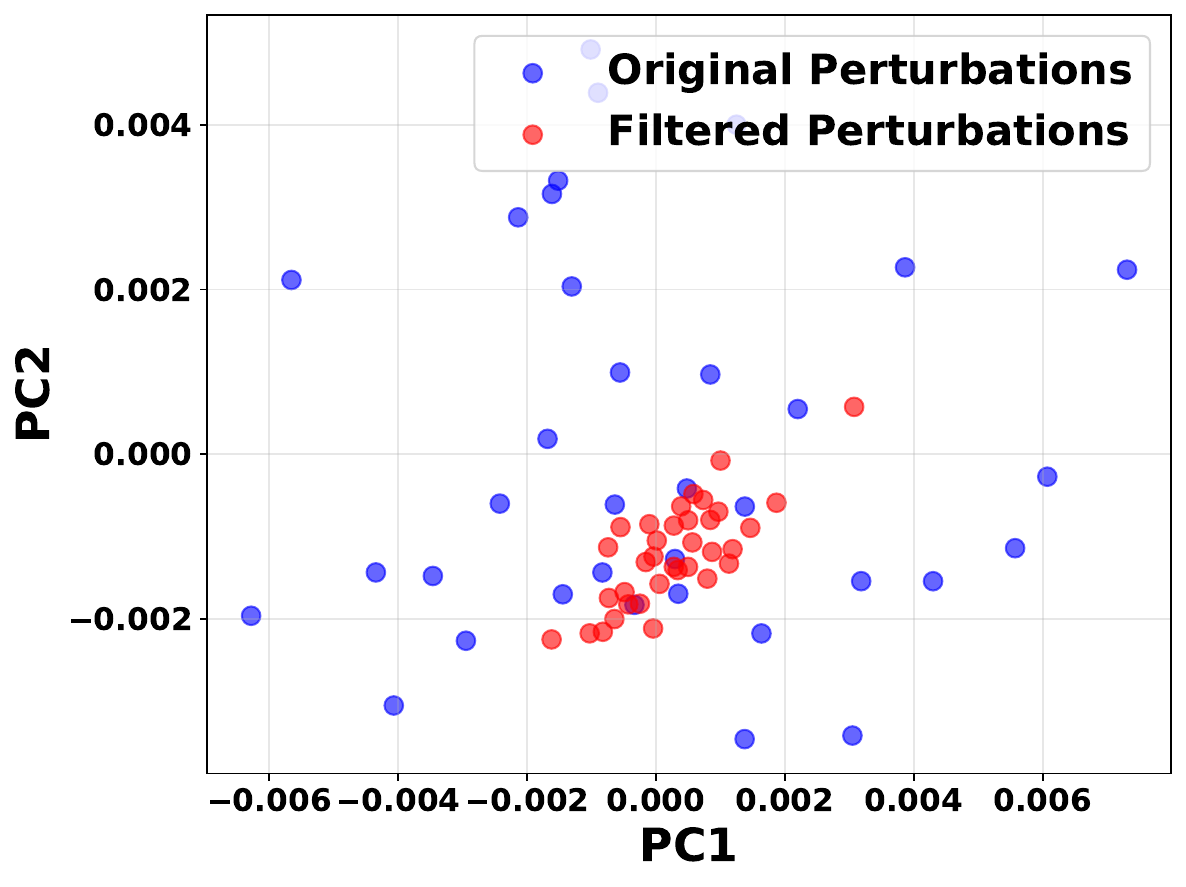}
        \centerline{\small (a) Client Perturbation Distribution}
        \label{fig:pca_p}
    \end{minipage}\hfill
    \begin{minipage}{0.31\textwidth}
        \centering
        \includegraphics[width=\linewidth]{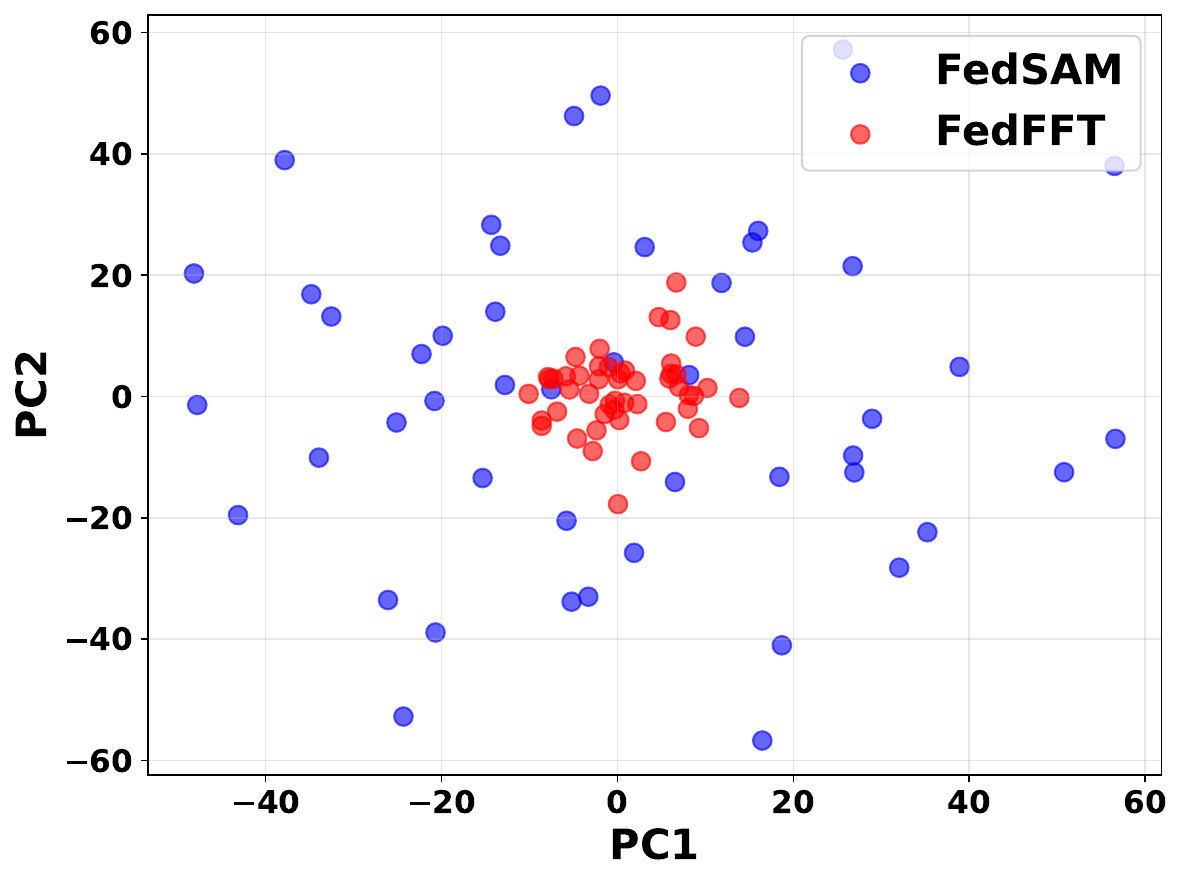}
        \centerline{\small (b) Model Feature Distribution}
        \label{fig:pca_feature}
    \end{minipage}\hfill
    \begin{minipage}{0.31\textwidth}
        \centering
        \includegraphics[width=\linewidth]{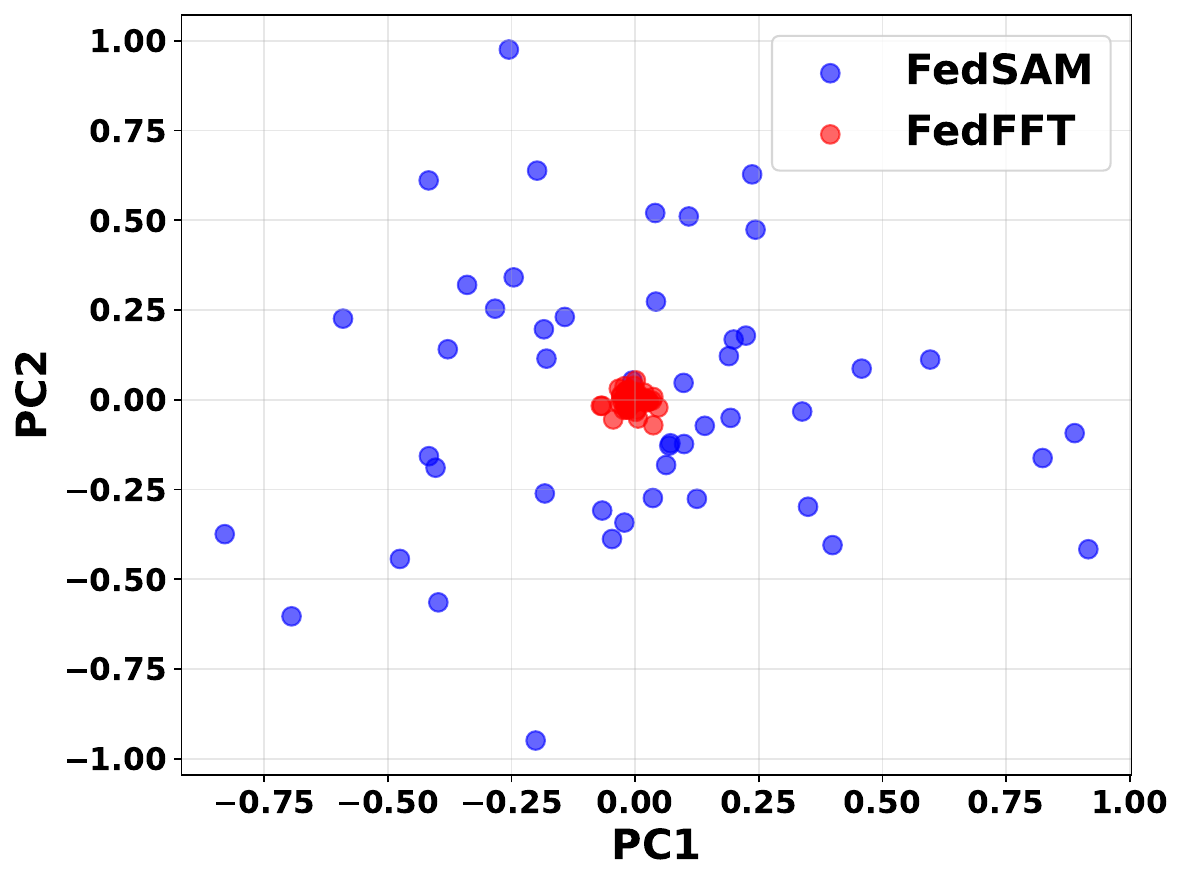}
        \centerline{\small (c) Client Model Parameters}
        \label{fig:pca_params}
    \end{minipage}

    \captionof{figure}{\small{Effect of low-frequency perturbation filtering on ResNet-18/CIFAR-10 ($\alpha=0.1$). The comparison between FedSAM (blue) and FedFFT (red) shows: (a) increased compactness of client perturbations (\texttt{conv1.weight}), leading to (b) more consistent model features (\texttt{layer4.1.gn2}) and (c) more aligned client model parameters (\texttt{conv1.weight}).}}
    \label{fig:pca_comparsion}
\end{minipage}
\vspace{1em} 

\begin{table}[t]
\centering
\caption{Test accuracy (\%) across different backbones on CIFAR-10 and CIFAR-100 with Dirichlet ($\alpha=0.6$). All methods use FedDyn as the base algorithm.}
\label{tab:arch_results}
\footnotesize 
\setlength{\tabcolsep}{3pt} 
\begin{tabular}{@{}llccccc@{}}
\toprule
Data & Method & ResNet18 & ResNet20 & DenseNet121 & ViT \\
\midrule
\multirow{4}{*}{CIFAR10} 
& FedSAM   & 82.29 & 88.82 & \underline{89.47} & 49.04 \\
& FedSMOO  & \underline{84.55} & \underline{89.86} & 88.72 & \underline{50.23} \\
& FedGloSS & 82.58 & 84.17 & 86.84 & 50.10 \\
& FedFFT   & \textbf{87.19} & \textbf{90.56} & \textbf{90.57} & \textbf{53.31} \\

\midrule
\multirow{4}{*}{CIFAR100} 
& FedSAM   & 53.70 & \underline{58.92} & \textbf{64.19} & 28.01 \\
& FedSMOO  & \underline{53.92} & 58.17 & \underline{63.74} & \underline{29.48} \\
& FedGloSS & 50.92 & 46.68 & 57.14 & 27.54 \\
& FedFFT   & \textbf{54.46} & \textbf{61.60} & 61.66 & \textbf{30.36} \\
\bottomrule
\end{tabular}
\end{table}

\section{Discussion}
\label{discussion}

\subsection{Ablation Study}

\subsubsection{Different filtering strategies.} 

\begin{table}[t]
    \centering
    \small 
    \setlength{\tabcolsep}{4pt} 
    \renewcommand{\arraystretch}{1.1} 
    
    \caption{Ablation of frequency-domain perturbation filtering strategies on CIFAR-10 using a ResNet-18 backbone.}
    \label{tab:filtering_methods_extended}
    
    \begin{tabular}{lcc}
    \toprule
    \textbf{Filtering Strategy} & $\alpha{=}0.6$ & $\alpha{=}0.1$ \\
    \midrule
    None (FedSAM)               & 81.91 & 74.92 \\
    High-frequency filtering    & 81.66 & 74.81 \\
    Random filtering            & 81.97 & 75.09 \\
    \emph{L2-norm rescaling}    & 81.88 & 74.79 \\
    \emph{Coordinate-wise clipping} & 80.13 & 75.19 \\
    \textbf{Low-frequency (Ours)} & \textbf{83.02} & \textbf{77.53} \\
    \bottomrule
    \end{tabular}
\end{table}

To validate our central hypothesis—that inter-client inconsistency is primarily
concentrated in the low-frequency spectrum—we compare our proposed
low-frequency filtering with several alternative perturbation-modification
strategies. In addition to the proposed low-frequency filtering, we compare against several alternative perturbation-modification strategies using the same filtering ratio of 0.01. Specifically, we evaluate high-frequency filtering, low-frequency filtering, and random filtering under an identical sparsity level. For random filtering, we first flatten the SAM perturbation and apply an rFFT transform; we then randomly select 1\% of the frequency coefficients and set them to zero. High-frequency filtering zeros out the top 1\% highest-frequency components, whereas low-frequency filtering zeros out the lowest 1\% frequency components. This controlled setup ensures that all methods modify perturbations with the same magnitude of frequency-domain sparsity, allowing for a fair comparison of their effects on inter-client consistency. Beyond high-frequency and random filtering baselines, we further examine two smoothing-style approaches applied to the SAM perturbation $\delta_k = \rho \cdot \nabla L_k(w_k)/\|\nabla L_k(w_k)\|_2$.

\emph{L2-norm rescaling} constrains the perturbation magnitude by projecting
$\delta_k$ back to the SAM radius whenever its $\ell_2$-norm exceeds $\rho$.
\emph{Coordinate-wise clipping} suppresses extreme perturbation values using a
threshold proportional to the median magnitude of~$\delta_k$. We evaluate all filtering strategies on CIFAR-10 under Dirichlet heterogeneity
with $\alpha = 0.6$ and $\alpha = 0.1$ using a ResNet-18 backbone. As shown in
Table~\ref{tab:filtering_methods_extended}, high-frequency filtering, random
filtering, L2-norm rescaling, and coordinate-wise clipping all produce negligible
improvements over the standard FedSAM baseline. In contrast, removing
low-frequency components consistently yields substantial and stable performance
gains across both heterogeneity levels. These results demonstrate that the effectiveness of FedFFT does not stem from
generic smoothing, coefficient truncation, or norm-based regularization, but
specifically from filtering the low-frequency spectral components of SAM
perturbations—the components exhibiting the largest cross-client disagreement
under heterogeneous federated settings.

\subsubsection{Impact of Filtering Ratio.} 

We explore the impact of varying the filtering ratio $r$ from $0.1\%$ to $8\%$ to highlight the flexibility and adaptability of hyperparameter tuning in our proposed FedFFT. As shown in Fig.~\ref{fig:filter_ratio}, we report the performance of different $r$ values in FedFFT, where we view FedSAM as baseline and use FedAvg, SCAFFOLD and FedDyn as the base algorithms, respectively. We can find that even with a small $r$ such as $0.1\%$, ours can consistently outperform its baseline. It proves that removing the discordant low-frequency components is beneficial for sharpness-aware federated optimization. Besides, high filtering ratios ($>7\%$) lead to gradual performance degradation, approaching or slightly falling below the baseline, which may remove useful information for optimization. Overall, these results confirm that our FedFFT can stably improve accuracy across all models when maintaining a reasonable filtering ratio. Based on our experiments, we recommend a safe range for $r$ between 0.5\% and 4\%.

\vspace{1em} 
\noindent\begin{minipage}{\textwidth}
    \centering
    \includegraphics[width=0.5\textwidth]{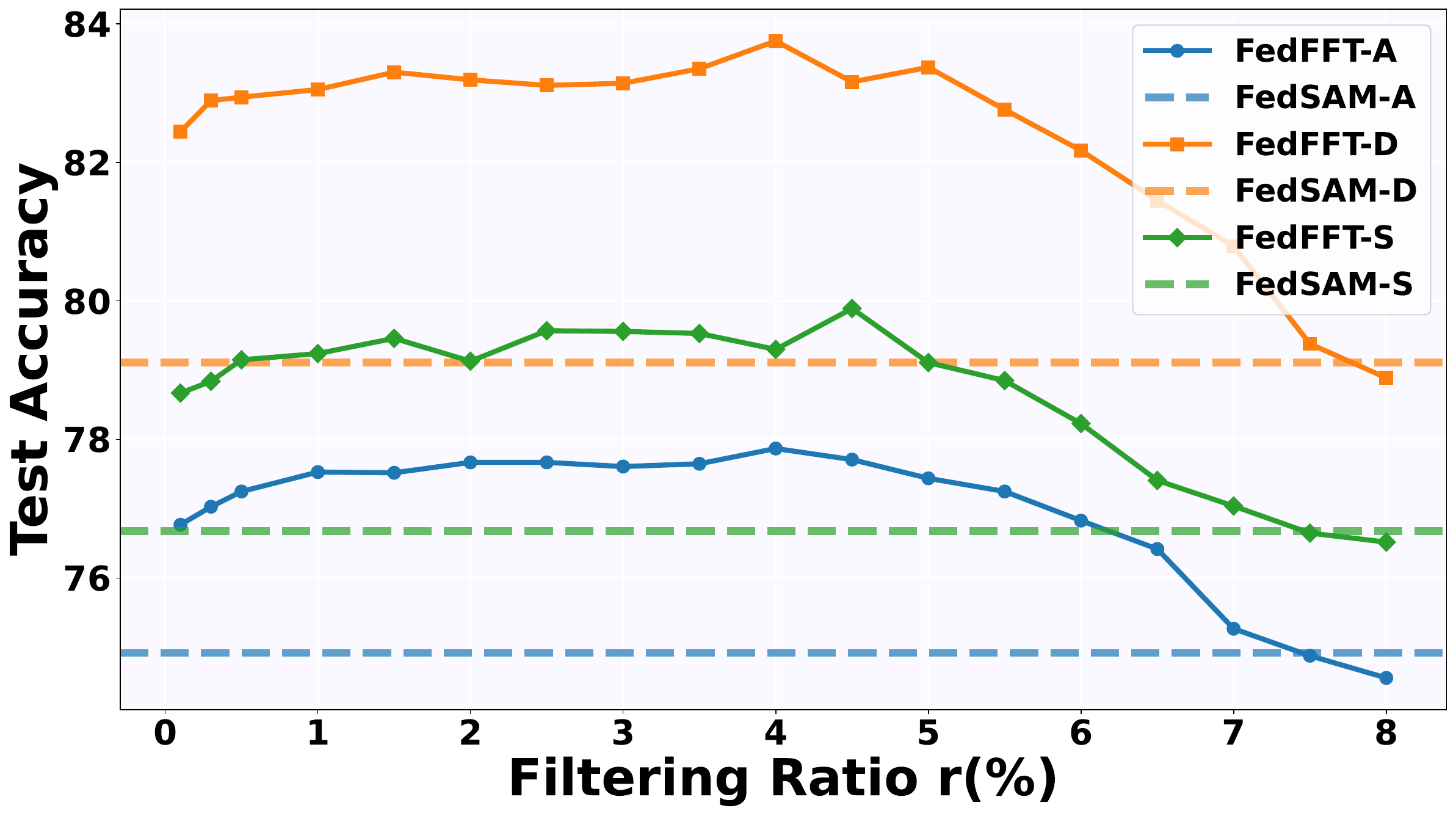}
    
    \captionof{figure}{\small Test accuracy with different filtering ratios on CIFAR-10 ($\alpha=0.1$). ``-A'', ``-S'' and ``-D'' denote using FedAvg, SCAFFOLD and FedDyn as the base algorithms.}
    \label{fig:filter_ratio}
\end{minipage}
\vspace{1em} 

\subsubsection{Robustness to FL Settings.}
We further conduct a systematic robustness analysis of FedFFT across three key federated learning hyperparameters: client activation rate, number of local training epochs, and total number of participating clients. The results, summarized in Fig. \ref{fig:performance_comparison}, show that FedFFT consistently and substantially outperforms FedSAM under all tested configurations, with its benefits becoming more pronounced as the learning conditions become more challenging.

In Fig. \ref{fig:performance_comparison}(a), we vary the activation rate from 5\% to 20\% while fixing the number of clients to 100 and local epochs to 5. Extremely low activation rates are known to intensify client drift and destabilize optimization, yet FedFFT retains a clear and stable advantage even at the most difficult setting of 5\% participation. As the activation rate increases, FedFFT continues to outperform FedSAM, though the performance gap narrows slightly—indicating that our spectral filtering is particularly beneficial under sparse-client participation regimes.

Fig. \ref{fig:performance_comparison}(b) examines the effect of local computation by varying the number of local epochs from 1 to 5, with activation rate fixed at 10\% and 100 clients. FedFFT exhibits the largest relative improvements at 1–2 epochs, conditions under which the optimization is more communication-constrained and convergence is harder to achieve. Although the margin decreases as local training deepens, FedFFT consistently maintains a clear lead across all settings, suggesting that its stabilization effect accelerates early-stage convergence while remaining beneficial even with more extensive local training.

Finally, Fig. \ref{fig:performance_comparison}(c) evaluates scalability by increasing the number of clients from 20 to 50, 80, and 100, while keeping local epochs at 5 and activation rate at 10\%. As the federation becomes larger—and thus more heterogeneous—FedFFT not only maintains but further widens its advantage over FedSAM. The largest improvements are observed when scaling from 20 to 100 clients, demonstrating that spectral filtering remains effective in large, heterogeneous networks.

Taken together, these results illustrate that the gains of FedFFT are not tied to a particular configuration, but instead hold broadly across a range of realistic and increasingly challenging FL conditions. Moreover, the trend that FedFFT delivers larger improvements when the FL environment becomes more difficult underscores its ability to mitigate drift, stabilize optimization, and enhance robustness in practical federated systems.

\subsubsection{Sensitivity to the Perturbation Radius.}

The perturbation radius~$\rho$ is an important hyperparameter in SAM-based federated optimization methods. In our main experiments, we follow common practice in the literature and set $\rho = 0.1$, consistent with prior studies such as FedSMOO and FedLESAM. To further examine how the choice of~$\rho$ influences performance, we conduct a sensitivity analysis for both FedSAM and our proposed FedFFT under a range of values: $\rho \in \{0.05,\, 0.10,\, 0.15,\, 0.20\}$.

Table~\ref{tab:perturb_radius_compare} summarizes the results on CIFAR-10 with Dirichlet $\alpha=0.1$ using a ResNet-18 backbone. We observe that FedFFT consistently outperforms FedSAM across all perturbation radii, and the performance gap increases as $\rho$ becomes larger. While FedSAM exhibits substantial degradation when the perturbation radius grows, FedFFT maintains stable accuracy, indicating significantly improved robustness. These findings suggest that the spectral filtering mechanism in FedFFT effectively stabilizes SAM-style perturbations under non-IID data distributions, making our method considerably less sensitive to the choice of~$\rho$.

\vspace{1em} 
\noindent\begin{minipage}{\textwidth}
    \centering
    
    \begin{minipage}{0.31\textwidth}
        \centering
        \includegraphics[width=\linewidth]{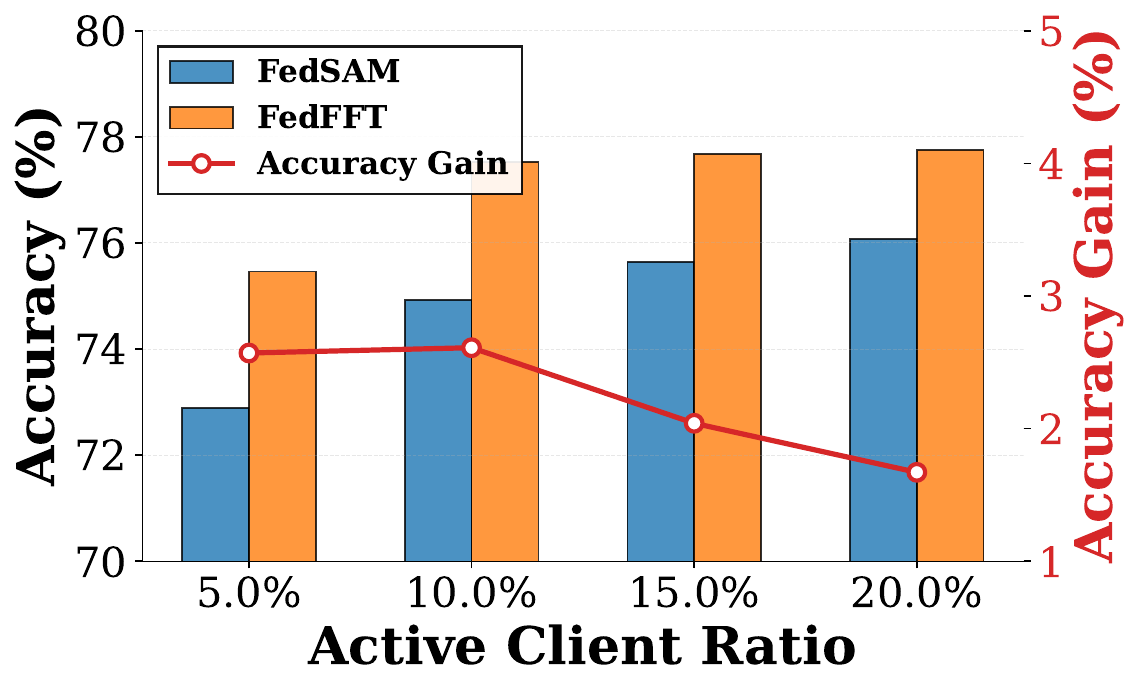}
        \centerline{\small (a) Active Ratio}
        \label{fig:active_ratio}
    \end{minipage}\hfill
    \begin{minipage}{0.31\textwidth}
        \centering
        \includegraphics[width=\linewidth]{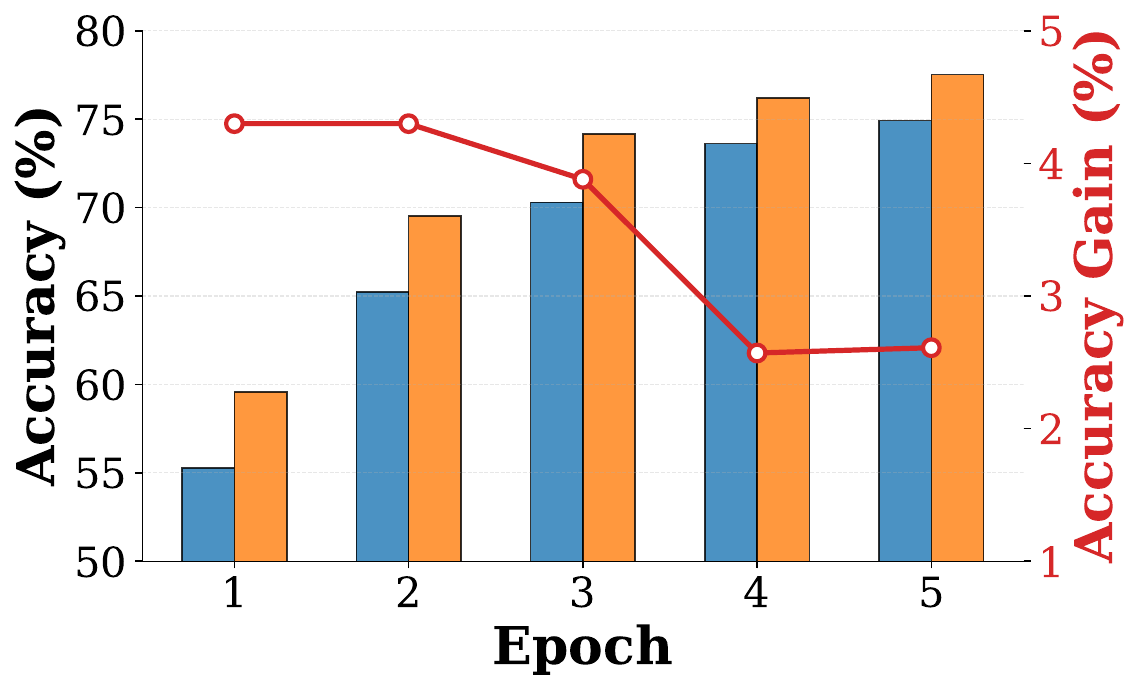}
        \centerline{\small (b) Local Epochs}
        \label{fig:epochs}
    \end{minipage}\hfill
    \begin{minipage}{0.31\textwidth}
        \centering
        \includegraphics[width=\linewidth]{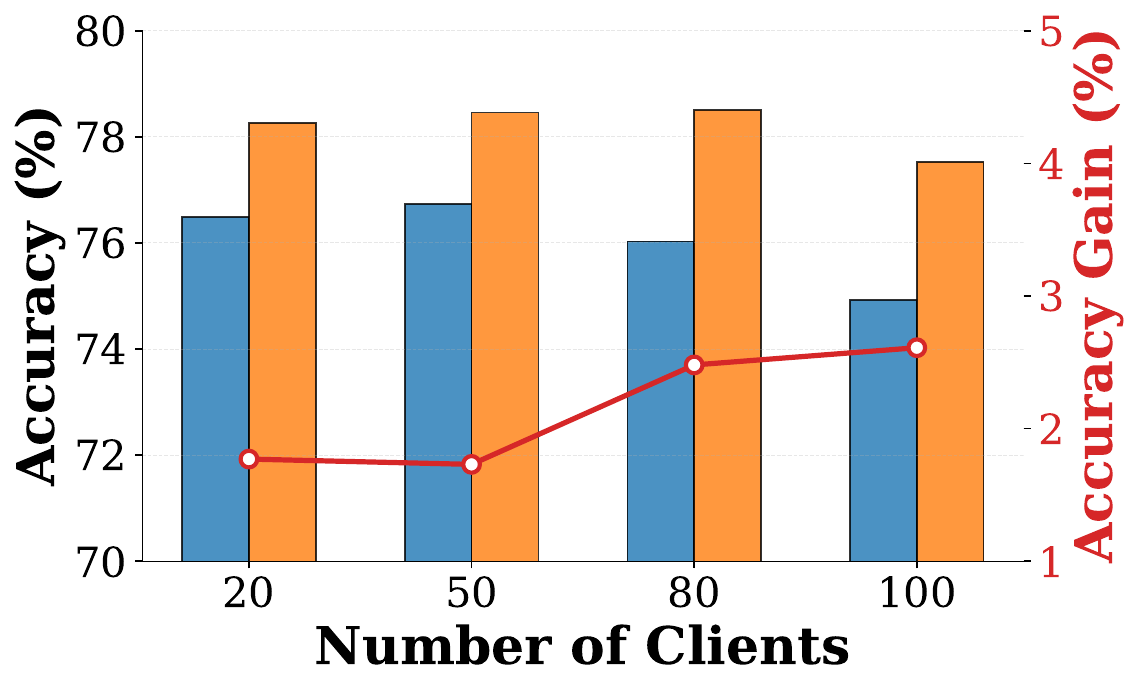}
        \centerline{\small (c) Client Numbers}
        \label{fig:client_nums}
    \end{minipage}

    \captionof{figure}{\small Comprehensive performance comparison between FedFFT and FedSAM across different experimental configurations on CIFAR-10 (Non-IID $\alpha=0.1$) using a ResNet-18 backbone.}
    \label{fig:performance_comparison}
\end{minipage}
\vspace{1em} 

\begin{table}[t]
\centering
\caption{Performance comparison between FedSAM and FedFFT under different perturbation radii $\rho$ on CIFAR-10 with Dirichlet $\alpha=0.1$ using ResNet-18.}
\label{tab:perturb_radius_compare}
\footnotesize 
\setlength{\tabcolsep}{3pt} 
\begin{tabular}{lcccc}
\toprule
\textbf{Method} & $\rho{=}0.05$ & $\rho{=}0.10$ & $\rho{=}0.15$ & $\rho{=}0.20$ \\
\midrule
FedSAM & 77.91 & 74.92 & 69.38 & 58.49 \\
FedFFT & 78.16 & 77.53 & 75.29 & 71.69 \\
\midrule
$\Delta$ (Improvement) & +0.25 & +2.61 & +5.91 & +13.20 \\
\bottomrule
\end{tabular}
\end{table}

\subsubsection{Behavior in Personalized Federated Learning Settings.}
Personalization is an important aspect in federated learning (FL), where methods must retain client-specific information while still benefiting from shared global knowledge. Here, we study how the proposed FedFFT behaves under personalized FL frameworks.
In FedPer, where each client maintains a private classification head while sharing the feature extractor. Since FedFFT suppresses high-frequency inconsistencies only within the shared representation—while the personalized head preserves client-unique high-frequency signals—the method does not hinder personalization and can even stabilize the learning of shared features.

To empirically validate this compatibility, we implemented FedPer \citep{fedper} with a ResNet-18 backbone on the FEMNIST dataset from the LEAF\citep{leaf} benchmark, a naturally federated dataset featuring substantial client heterogeneity. For each client, we replaced the local optimizer in FedPer with  
(i) a SAM optimizer and  
(ii) our FFT-based SAM optimizer,  
while keeping all other components identical.

As shown in Table~\ref{tab:combined_results}, FedFFT consistently yields higher accuracy compared to SAM and exhibits smoother convergence behavior. These results confirm that FedFFT effectively preserves client-specific signals while improving the stability of shared representation learning.

\begin{table}[htbp]
\centering
\footnotesize
\caption{Comparison between SAM and FFT-SAM optimizers and test accuracy on CIFAR-10/100 with Dirichlet $\alpha=0.1$.}
\label{tab:combined_results}

\begin{minipage}[t]{0.45\textwidth}
\centering
\subcaption*{FEMNIST with ResNet-18 in Personalized FL}
\setlength{\tabcolsep}{4pt}
\begin{tabular}{lc}
\toprule
\textbf{Method} & \textbf{Accuracy} \\
\midrule
FedPer  & 63.76 \\
FedPer + SAM & 65.49 \\
FedPer + \textbf{FFT-SAM (ours)} & \textbf{67.12} \\
\bottomrule
\end{tabular}
\end{minipage}
\hfill
\begin{minipage}[t]{0.45\textwidth}
\centering
\subcaption*{Combined with FedAWA}
\setlength{\tabcolsep}{4pt}
\begin{tabular}{lcc}
\toprule
\textbf{Method} & \textbf{CIFAR-10} & \textbf{CIFAR-100} \\
\midrule
FedAWA  & 47.20 & 33.40 \\
FedAWA+FFT & \textbf{50.16} & \textbf{34.04} \\
\bottomrule
\end{tabular}
\end{minipage}

\medskip

\end{table}

\subsubsection{Combine with other methods}
To further validate the universality and compatibility of FedFFT, we integrate it with FedAWA \cite{fedawa}, a representative method that improves federated optimization by adaptively learning aggregation weights based on client contributions. This experiment aims to demonstrate that our frequency-domain perturbation filtering is orthogonal to existing FL improvements and can be seamlessly incorporated without modifying their algorithmic pipelines.

Specifically, we adopt the standard FedAvg training framework and implement the SAM optimizer on all participating clients. FedFFT is then applied as a drop-in replacement for the original SAM perturbation computation, while FedAWA remains responsible for determining aggregation weights on the server side. This setup ensures that any performance changes can be attributed solely to the effect of frequency-domain filtering rather than confounding factors introduced by altering the aggregation strategy.

We evaluate this combined method using the ResNet-20 model under heterogeneous client data distributions. As shown in {Table}~\ref{tab:combined_results}, integrating FedFFT consistently enhances the performance of FedAWA across multiple metrics. This demonstrates that our approach can serve as a general-purpose enhancement module that strengthens optimization stability and robustness, regardless of the underlying federated learning algorithm.

\section{Conclusion}
\label{sec:conclusion}

In this work, we introduce a novel frequency-domain perspective to address divergent perturbations in sharpness-aware federated learning. We identify that inter-client disagreements are predominantly a low-frequency phenomenon and accordingly propose FedFFT, a lightweight filtering method to suppress these discordant components. Extensive experiments validate that FedFFT consistently outperforms state-of-the-art methods, particularly in highly non-IID settings, by converging to visibly flatter and wider global minima. This result not only explains the superior generalization and communication efficiency of our method but also establishes spectral analysis as a powerful new tool for designing robust federated optimization algorithms.

\section*{Declaration of Generative AI and AI-assisted Technologies in the Writing Process}

During the preparation of this work, the author(s) used generative AI tools, including ChatGPT, in order to improve the clarity, grammar, and overall readability of the manuscript. After using these tools, the author(s) carefully reviewed and edited the content as needed and take full responsibility for the content of the published article.

\section{CRediT authorship contribution statement.}

Liyang Yuan: Conceptualization, Methodology, Software, Validation, Formal analysis, Investigation, Writing - original draft, Visualization. Dandan Guo: Conceptualization, Resources, Writing - review \& editing, Supervision, Project administration, Funding acquisition. Yibo Yang: Conceptualization, Methodology, Writing - review \& editing, Supervision.

\bibliographystyle{elsarticle-num}
\bibliography{refs}

\end{document}